DEEP LEARNING MIRACULOUS YEAR 1990-91

JÜRGEN SCHMIDHUBER 2020

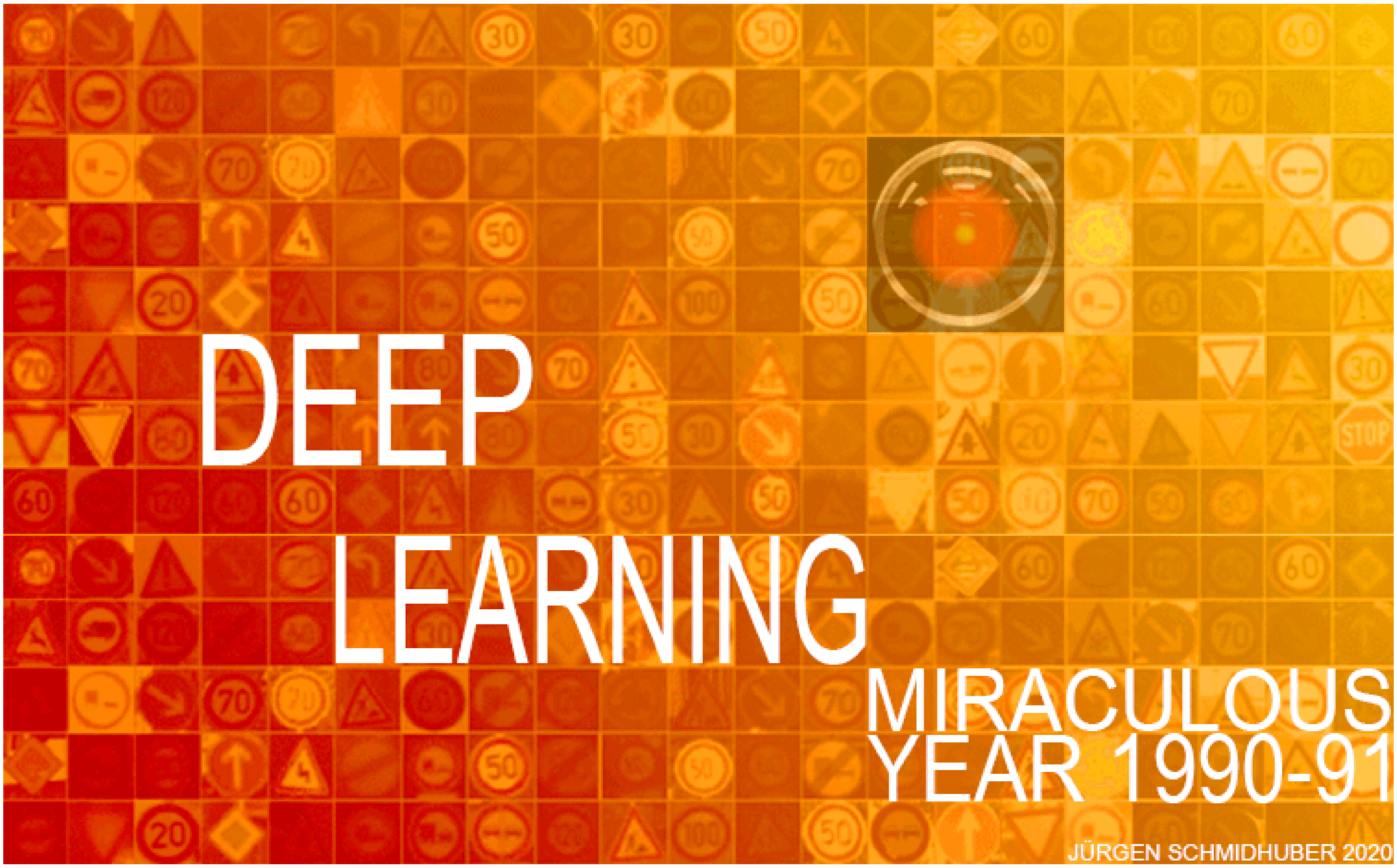

Jürgen Schmidhuber (2019, updated '20, '21, '22, '25, '26)
Pronounce: You_again Shmidhoobuh
Preprint arxiv:2005.05744

AI Blog
@SchmidhuberAI
Inaugural tweet: 4 Oct 2019

---

# Deep Learning: Our Miraculous Year 1990-1991

---

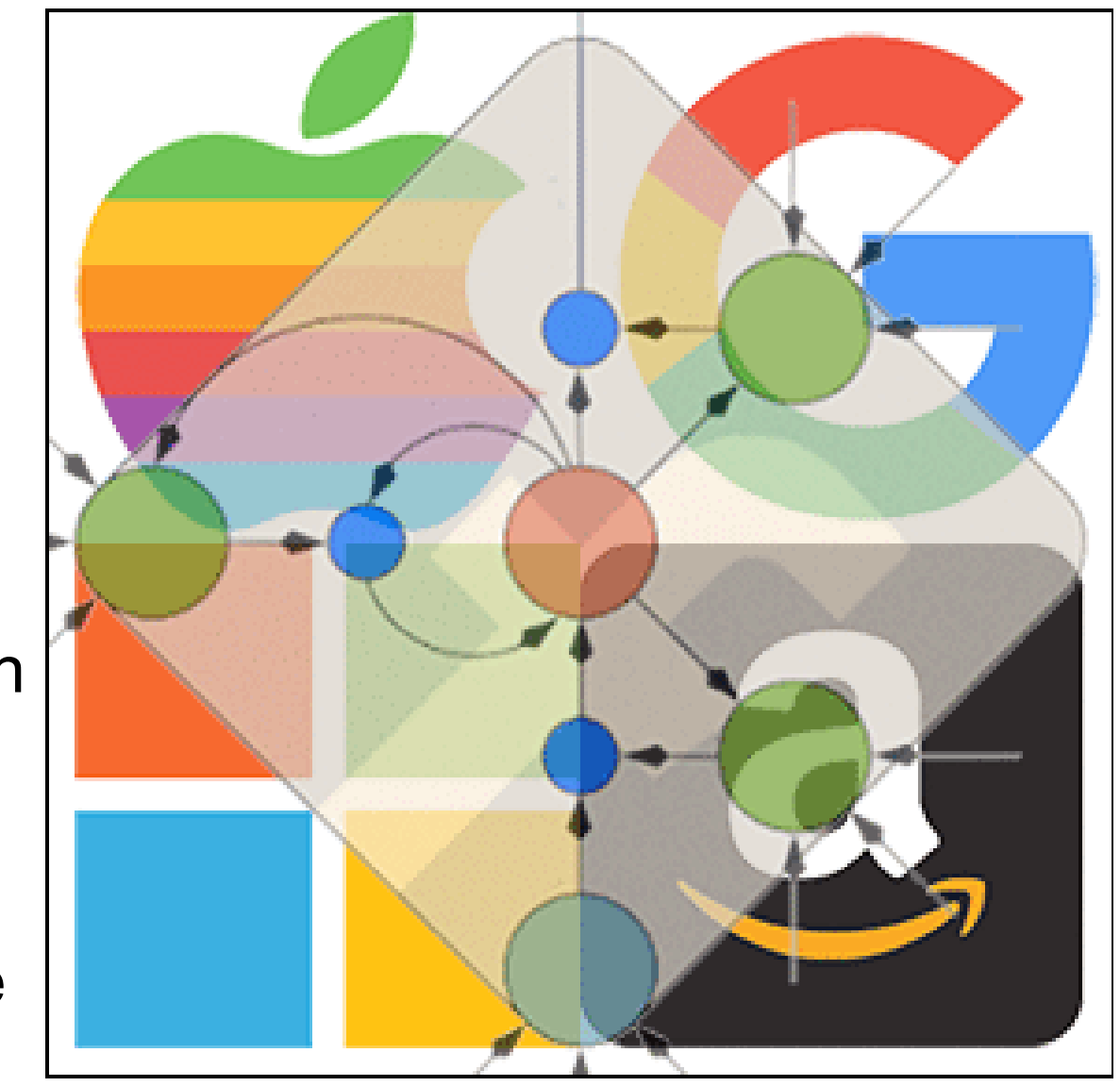

**Abstract.** The Deep Learning Artificial Neural Networks (NNs) of our team have revolutionised Machine Learning & AI. Many of the basic ideas behind this revolution were published within the 12 months of our *"Annus Mirabilis"* 1990-1991 at our lab in TU Munich. Back then, few people were interested. But a quarter century later, NNs based on our *"Miraculous Year"* were on over 3 billion devices, and used many billions of times per day, consuming a significant fraction of the world's compute. In particular, in 1990-91, we laid foundations of Generative AI, publishing principles of (1) Generative Adversarial Networks for Artificial Curiosity and Creativity (now used for deepfakes), (2) Transformers (the T in ChatGPT—see the 1991 Unnormalized Linear Transformer), (3) Pre-training for deep NNs (see the P in ChatGPT), (4) NN distillation (key for DeepSeek), and (5) recurrent World Models for Reinforcement Learning and Planning in partially observable environments. The year 1991 also marks the emergence of the defining features of (6) LSTM, the most cited AI paper of the 20th century (based on deep residual learning and constant error flow through residual NN connections), and (7) the most cited paper of the 21st century, based on our LSTM-inspired Highway Net that was 10 times deeper than previous feedforward NNs. As of 2025, the two most frequently cited scientific articles of all time (with the most Google Scholar citations within 3 years—manuals excluded) are both directly based on our 1991 work.[MOST26]

The following summary of what happened in 1990-91 both contains some high-level context which is accessible by a general audience, but also references to original sources for NN experts. I also mention selected later work which further developed the ideas of 1990-91 (at TU Munich, the Swiss AI Lab IDSIA, and other places), as well as related work by others.

# Background on Deep Learning in Neural Nets (NNs)

The human brain has on the order of 100 billion neurons, each connected to 10,000 other neurons on average. Some are input neurons that feed the rest with data (sound, vision, tactile, pain, hunger). Others are output neurons that control muscles. Most neurons are hidden in between, where thinking takes place. Your brain apparently learns by changing the strengths or weights of the connections, which determine how strongly neurons influence each other, and which seem to encode all your lifelong experience. Similar for our *artifical* neural networks (NNs), which learn better than previous methods to recognize speech or handwriting or video, minimize pain, maximize pleasure, drive cars, etc.[DL1-4][DLH]

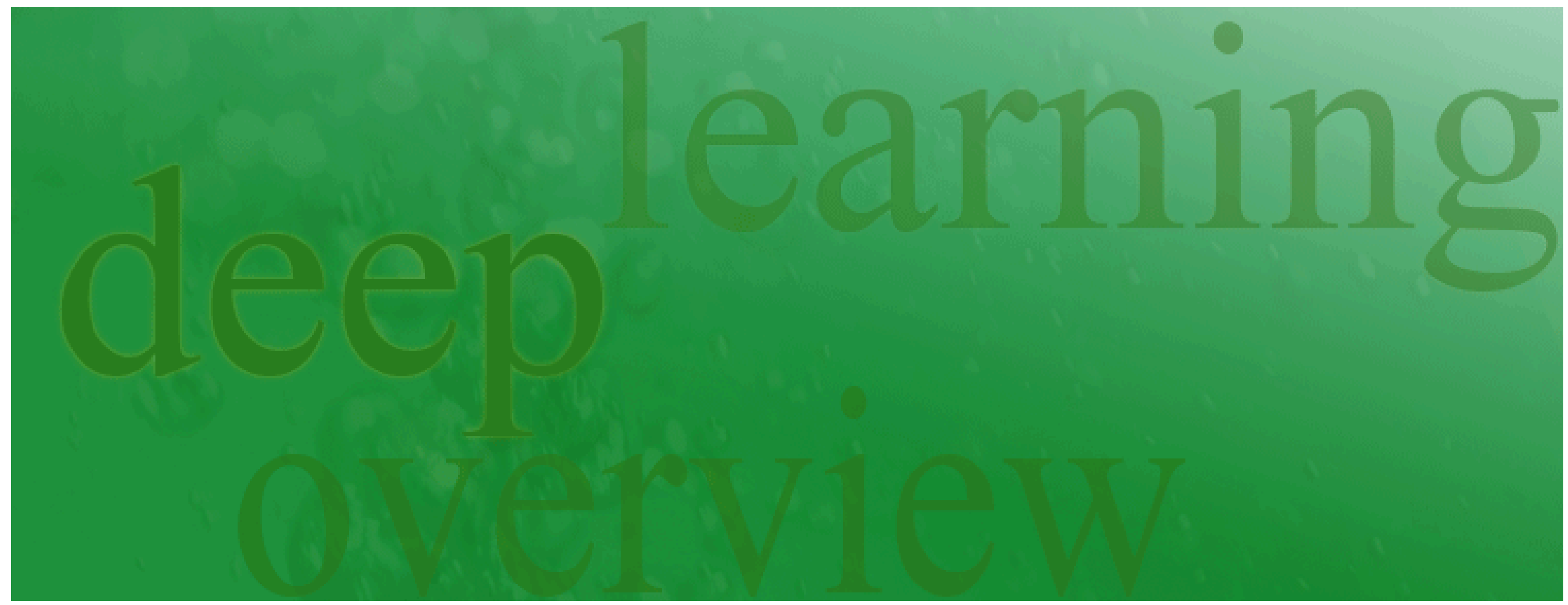

Most current commercial applications focus on supervised learning to make NNs imitate human teachers.[DL1-4] In the course of many trials, Seppo Linnainmaa's gradient-computing algorithm of 1970,[BP1][DLH] today often called backpropagation or the reverse mode of automatic differentiation,[BP4] is used to incrementally weaken certain NN connections and strengthen others, such that the NN behaves more and more like the teacher.[BPA-C][BP2][HIN][T22][DLP][NOB]

Today's most powerful NNs tend to be very deep, that is, they have many layers of neurons or many subsequent computational stages. In the 1980s, however, gradient-based training did not work well for deep NNs, only for shallow ones.[DL1-2]

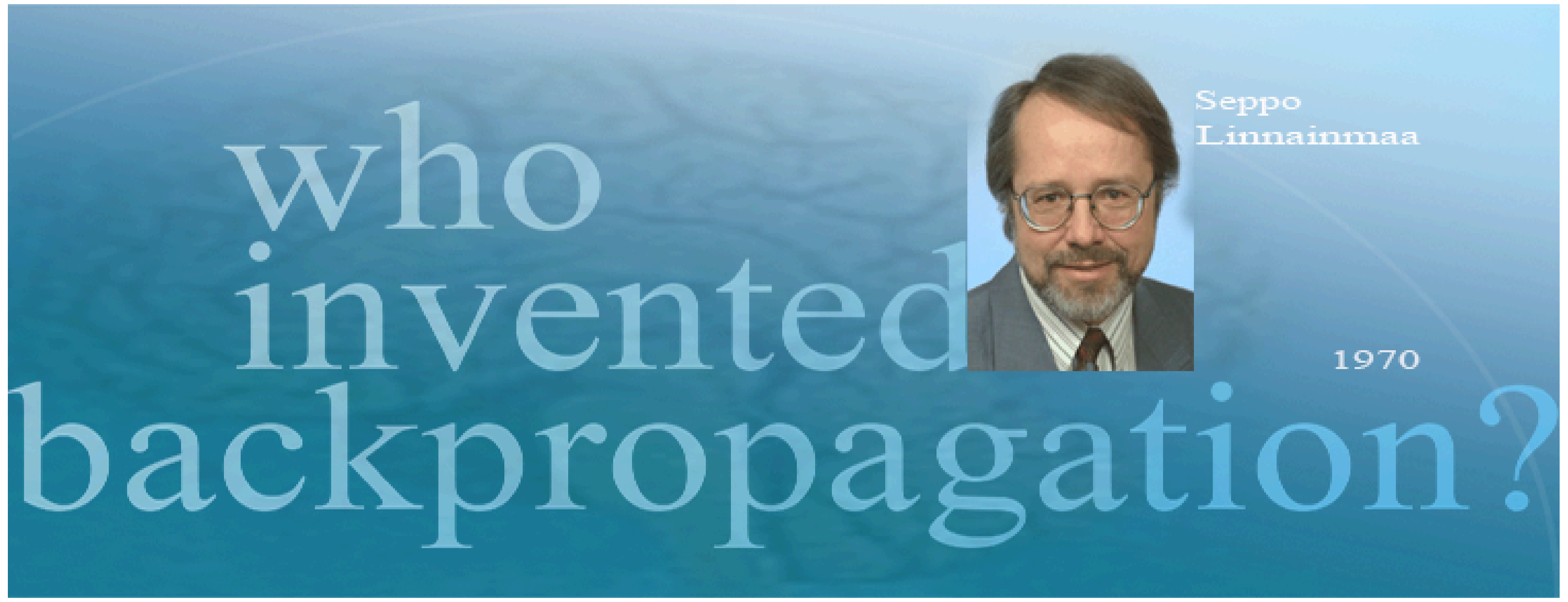

This *Deep Learning Problem* was most obvious for *recurrent* NNs (RNNs). RNN architectures have been studied since the 1920s.[L20][I24-I25][K41][MC43][W45][K56][AMH1-2][NOB] Like the human brain, but unlike the more limited *feedforward* NNs (FNNs), RNNs have feedback connections. This makes RNNs powerful, general purpose, parallel-sequential computers that can process input sequences of arbitrary length (think of speech data or videos). RNNs can in principle implement any program that can run on your laptop or any other computer in existence. If we want to build an *Artificial General Intelligence* (AGI), then its underlying computational substrate must be something more like an RNN than an FNN as FNNs are fundamentally insufficient; RNNs are to FNNs as general computers are to pocket calculators.

In particular, unlike FNNs, RNNs can in principle deal with problems of arbitrary depth.[DL1] Early RNNs of the 1980s, however, failed to learn deep problems in practice. From early in my career, I wanted to overcome this drawback, to achieve RNN-based *"general purpose Deep Learning"* or *"general Deep Learning."*

# 1. First Very Deep NNs, Based on Unsupervised or Self-Supervised Pre-Training (1991, see the P in ChatGPT)

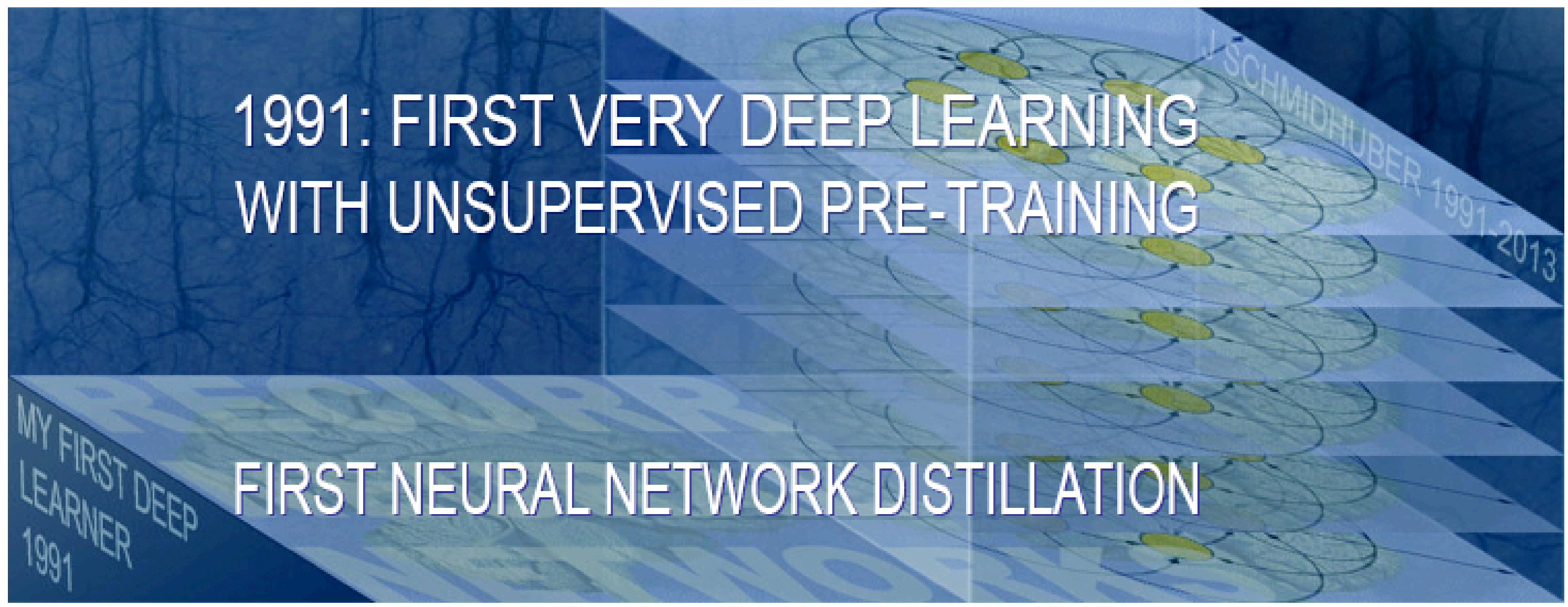

My first idea to overcome the *Deep Learning Problem* mentioned above was to facilitate supervised learning in deep RNNs by unsupervised pre-training of a hierarchical stack of RNNs (1991, see the P in ChatGPT), to obtain a first *"Very Deep Learner"* called the *Neural Sequence Chunker*[UN0] or Neural History Compressor.[UN1] Each higher level minimizes the description length (or negative log probability) of the data representation in the level below, using the *Predictive Coding* trick: try to predict the next input in the incoming data stream, given the previous inputs, and update neural activations only in case of *unpredictable data*, thus storing only what's not yet known. In other words, the chunker learns to compress the data stream such that the *Deep Learning Problem* becomes less severe, and can be solved by standard backpropagation. Although computers back then were about a million times slower per dollar than today, by 1993, my method was able to solve previously unsolvable "Very Deep Learning" tasks of depth > 1000[UN2] (requiring more than 1,000 subsequent computational stages—the more such stages, the deeper the learning). In 1993, we also published a *continuous* version of the Neural History Compressor.[UN3]

To my knowledge, the Sequence Chunker[UN0] also was the first system made of RNNs operating on different (self-organizing) time scales.[UN][DLP] (But I also had a way of distilling all those RNNs down into a single deep RNN operating on a single time scale—see Sec. 2.) A few years later, others also started publishing on multi-time scale RNNs[HB96][DLP] (see also the *Clockwork* RNN[CW]).

More than a decade after this work,[UN1] a similar method for more limited *feedforward* NNs (FNNs) was published, facilitating supervised learning by unsupervised pre-training of stacks of FNNs called *Deep Belief Networks* (DBNs).[UN4][DLP][NOB] The 2006 justification was essentially the one I used in the early 1990s for my RNN stack: each higher level tries to reduce the description length (or negative log probability) of the data representation in the level below.[HIN]

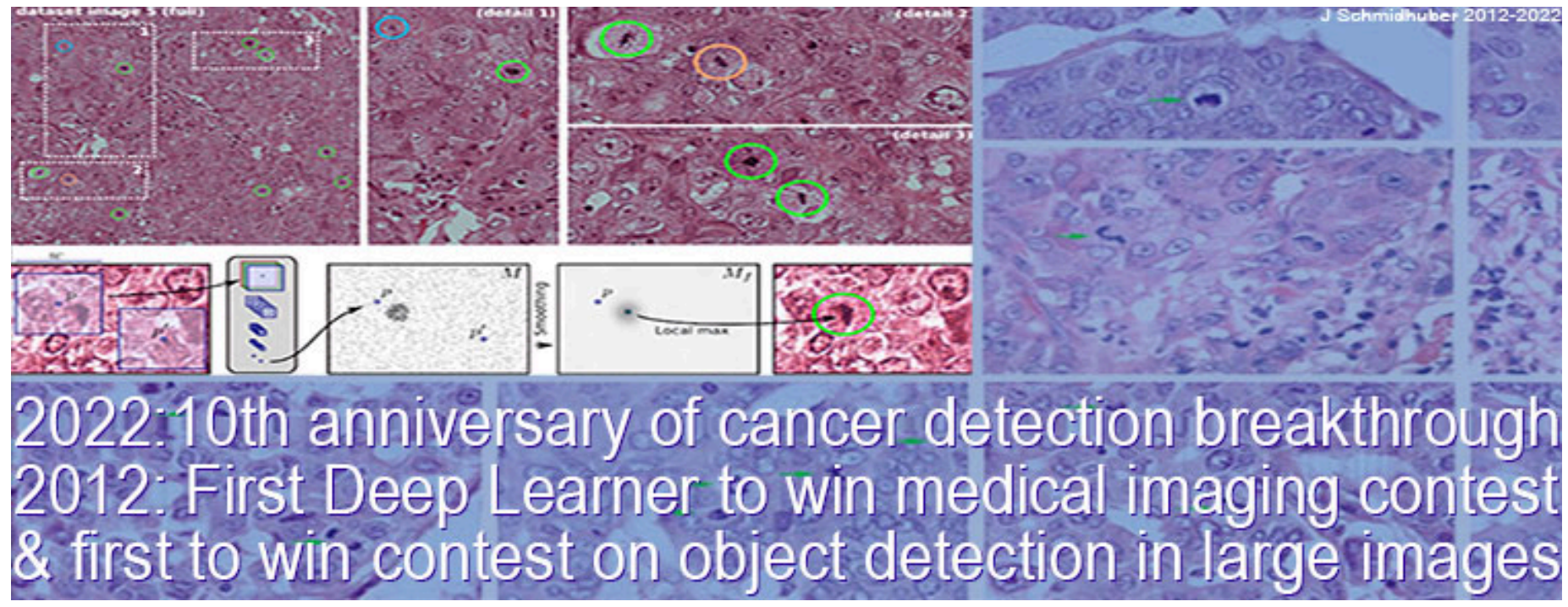

Soon after the unsupervised pre-training-based Very Deep Learner above, the Deep Learning Problem (see Sec. 3) was also overcome through our *purely supervised* LSTM (Sec. 4). Much later, between 2006 and 2011, my lab also drove a very similar shift from unsupervised pre-training to pure supervised learning, two decades after our *Miraculous Year*, this time for the less general *feedforward* NNs (FNNs) rather than *recurrent* NNs (RNNs), with revolutionary applications to cancer detection and many other problems. See Sec. 19 for more on this.

Of course, Deep Learning in feedforward NNs started much earlier, with Ivakhnenko & Lapa, who published the first general, working learning algorithms for deep multilayer perceptrons with arbitrarily many layers back in 1965.[DEEP1][NOB][DLH] For example, Ivakhnenko's paper from 1971[DEEP2] already described a Deep Learning net with 8 layers, trained by a highly cited method still popular in the new millennium.[DL2] But unlike the deep FNNs of Ivakhnenko and his successors of the 1970s and 80s, our deep RNNs had general purpose parallel-sequential computational architectures.[UN-UN3] By the early 1990s, most NN research was still limited to rather shallow nets with fewer than 10 subsequent computational stages, while our methods already enabled over 1,000 such stages. In this sense we were the ones who made NNs really deep, especially RNNs, the deepest and most powerful nets of them all. See also: who invented deep learning?[WHO5]

## 2. Compressing/Collapsing/Distilling an NN into Another NN (1991; key for DeepSeek 2025)

My above-mentioned paper on the Neural History Compressor (see Sec. 1) also introduced a way of compressing the network hierarchy (whose higher levels are typically running on much slower self-organising time scales than lower levels) into a *single* deep RNN[UN1] which thus learned to solve very deep problems despite the obstacles mentioned in Sec. 0. This is described in Section 4 of the paper on the "conscious" chunker and a "subconscious" automatiser,[UN1] which introduced a general principle for transferring the knowledge of one NN to another. Suppose a teacher NN has learned to predict (conditional expectations of) data, given other data. Its knowledge can be compressed into a student NN, by training the student

NN to imitate the behavior of the teacher NN (while also re-training the student NN on previously learned skills such that it does not forget them).

I called this "collapsing" or "compressing" the behavior of one net into another. Today, this is widely used, and also called "distilling"[DIST2][HIN][DLP] or "cloning" the behavior of a teacher net into a student net. See also: who invented knowledge distillation with artificial neural networks?[WHO9]

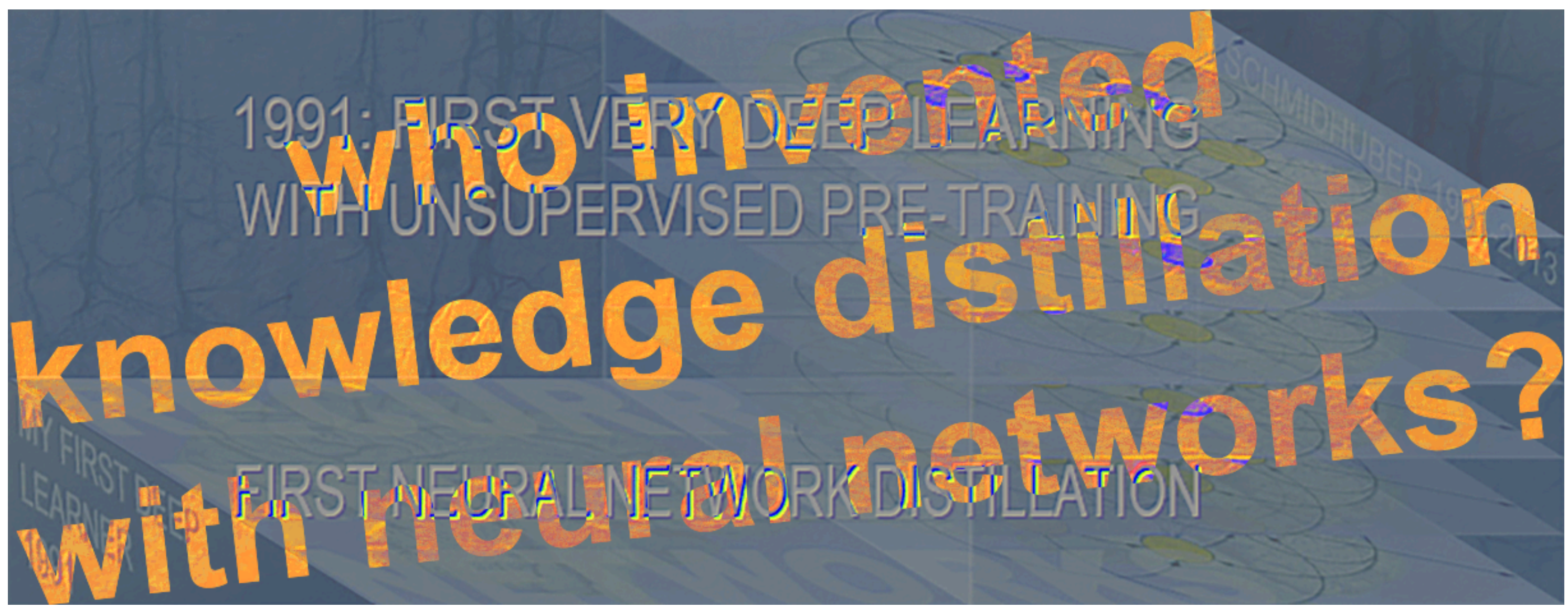

In January 2025, the **DeepSeek** "Sputnik"[DS1] wiped out a trillion USD from the stock market. DeepSeek-R1[DS1] used elements of my 2015 reinforcement learning (RL) prompt engineer[PLAN4] and its 2018 refinement[PLAN5] which collapses the 2015 RL machine and its world model[PLAN4] into a single net through the neural net distillation procedure of 1991[UN0-3][UN][DLP]: a distilled chain of thought system. See the popular tweet of 31 Jan 2025.

# 3. The Fundamental Deep Learning Problem: Vanishing / Exploding Gradients (1991)

The background section Sec. 0 pointed out that Deep Learning is hard. But why is it hard? A main reason is what I like to call the Fundamental Deep Learning Problem identified and analyzed in 1991 by my first student Sepp Hochreiter in his diploma thesis.[VAN1]

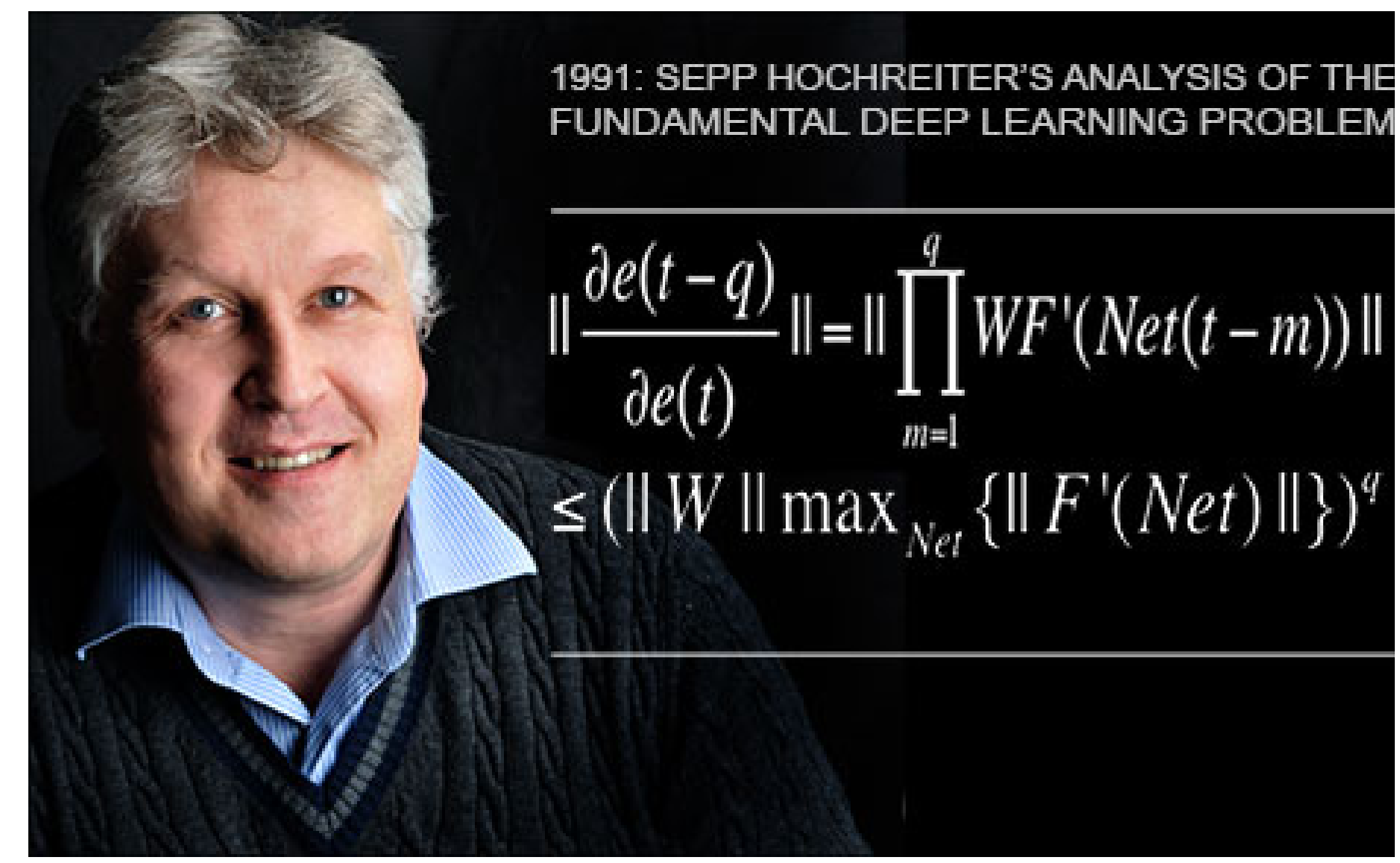

As a part of his thesis, Sepp implemented the Neural History Compressor above (see Sec. 1) and other RNN-based systems (see Sec. 11). However, he did much more: His work formally showed that deep NNs suffer from the now famous problem of vanishing or exploding gradients: in typical deep or recurrent networks, back-propagated error signals either shrink rapidly, or grow out of bounds. In both cases, learning fails. This analysis led to basic principles of what's now called LSTM (see Sec. 4).

Note that Sepp's thesis identified those problems of backpropagation in deep NNs two decades after another student with a similar first name (Seppo Linnainmaa) published modern backpropagation or the reverse mode of automatic differentiation in his own thesis of 1970.[BP1]

In 1994, others published results[VAN2] essentially identical to the 1991 vanishing gradient results of Sepp.[VAN1] Even after a common publication[VAN3] the first author of reference[VAN2] published papers[VAN4] that cited only their own 1994 paper but not Sepp's original work.[DLP]

# 4. Long Short-Term Memory Recurrent Networks: Supervised Very Deep Learning / Residual Connections

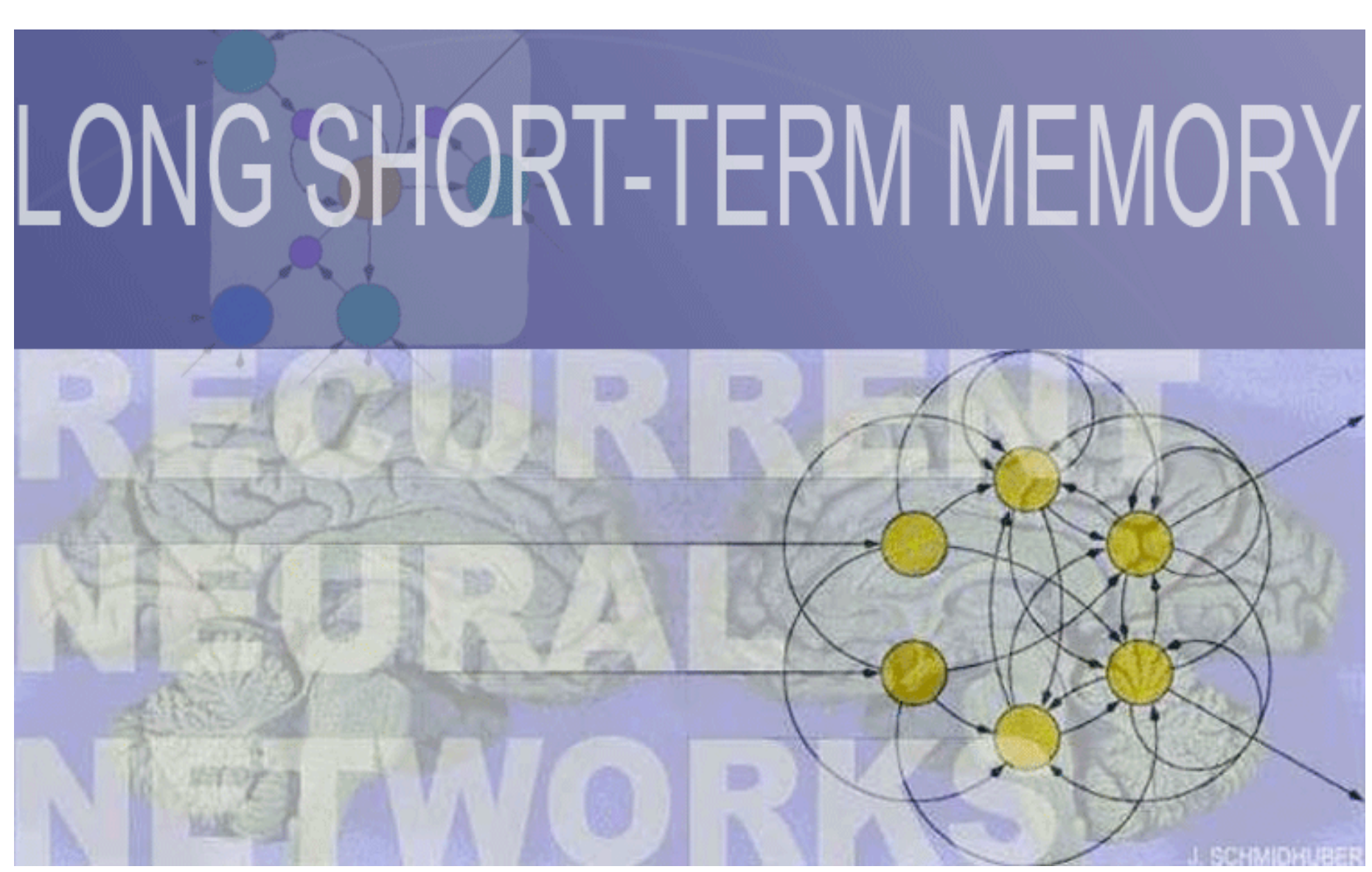

The Long Short-Term Memory (LSTM) recurrent neural network[LSTM1-6] overcomes the Fundamental Deep Learning Problem identified by Sepp in his above-mentioned 1991 diploma thesis[VAN1] (see Sec. 3), which I consider one of the most important documents in the history of machine learning. It also provided essential insights for overcoming the problem, through basic principles (such as *constant error flow through what's now called "residual connections"*) of what we called LSTM in a tech report of 1995.[LSTM0] This led to lots of follow-up work described below. See also: who invented deep residual learning?[WHO11]

In 2020 we celebrated the quarter-century anniversary of LSTM's first failure to pass peer review. After the main peer-reviewed publication in 1997[LSTM1] (now the most cited article in the history of *Neural Computation*[25y97]), LSTM and its training procedures were further improved on my Swiss LSTM grants at IDSIA through the work of my later students Felix Gers, Alex Graves, and others. A milestone was the *"vanilla LSTM architecture"* with forget gate[LSTM2a][LSTM2]—the LSTM variant of 1999-2000 that everybody is using today, e.g., in Google's Tensorflow.

Alex was lead author of our first successful application of LSTM to speech (2004).[LSTM10] 2005 saw the first publication of LSTM with full backpropagation through time and of bi-directional LSTM[LSTM3] (now widely used). Another milestone of 2006 was the training method *"Connectionist Temporal Classification"* or CTC[CTC] for simultaneous alignment and recognition of sequences. Our team successfully applied CTC-trained LSTM to speech in 2007[LSTM4] (also with hierarchical LSTM stacks[LSTM14]). This was the first superior end-to-end neural speech recognition. It was very different from hybrid methods since the late 1980s which combined NNs and traditional approaches such as Hidden Markov Models (HMMs).[BW][BRI][BOU][HYB12][DLP] In 2015, the CTC-LSTM combination dramatically improved Google's speech recognition on the Android smartphones.[GSR15][DL4][DLH]

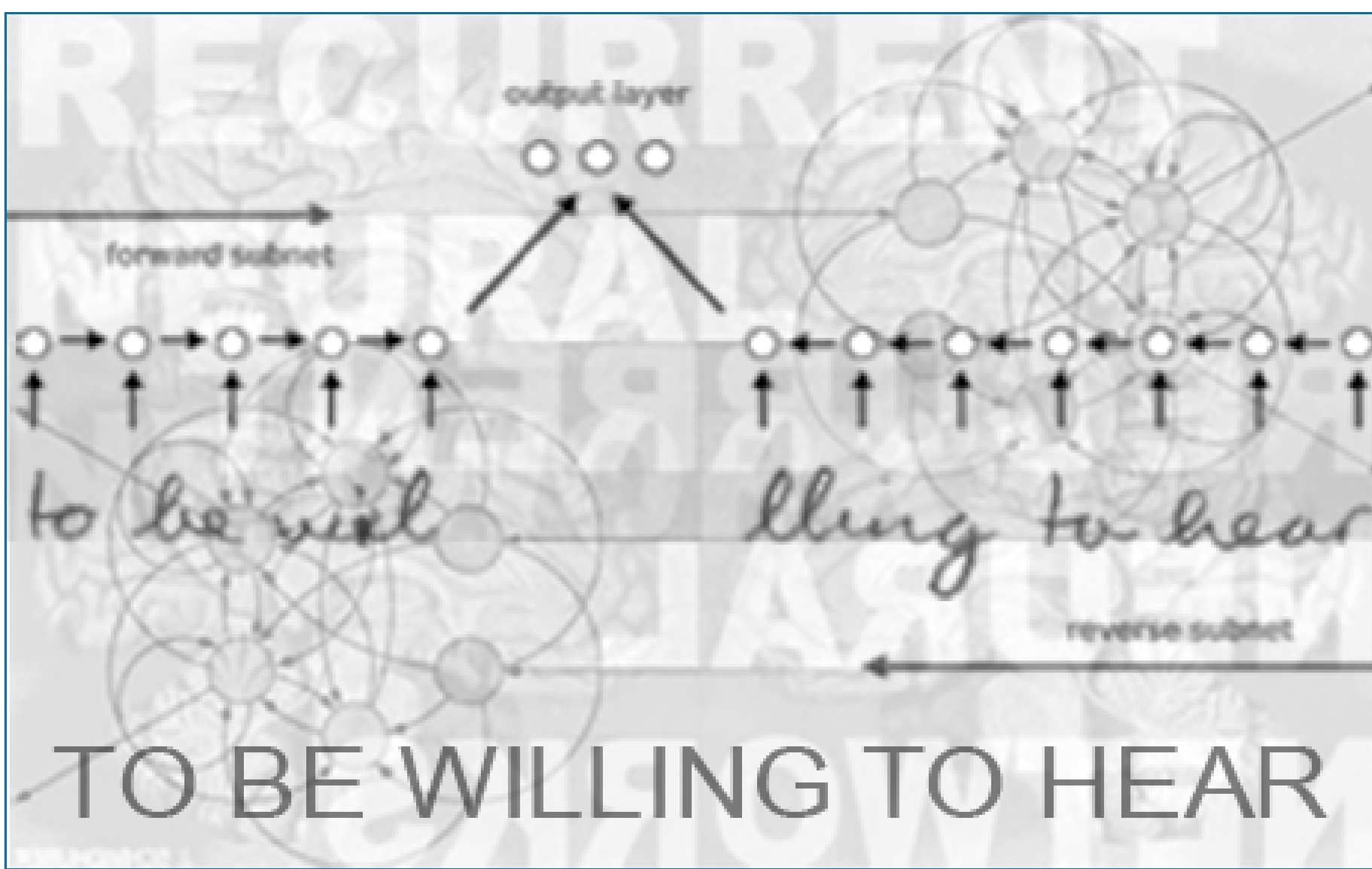

The first superior end-to-end neural machine translation was also based on our LSTM. In 1995, we already had an excellent neural probabilistic text model.[SNT] In the early 2000s, we showed how LSTM can learn languages unlearnable by traditional models such as Hidden Markov Models.[LSTM13] This took a while to sink in, and compute still had to get 1000 times cheaper, but by 2016-17, both Google Translate[WU][GT16] and Facebook Translate[FB17] were based on two connected LSTMs, one for the incoming text, one for the outgoing translation,[S2Sa,b,c,d] much better than what they had before.[DL4]

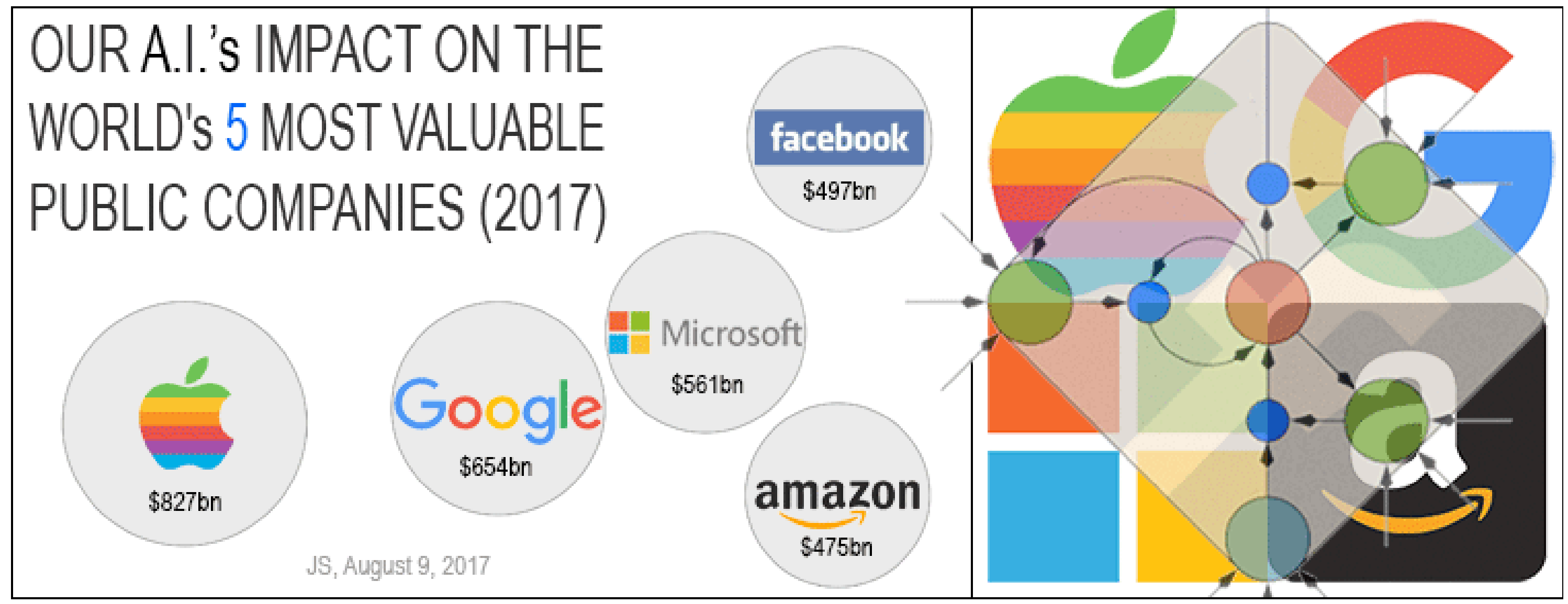

In 2009, my PhD student Justin Bayer was lead author of a system that automatically designed LSTM-like architectures outperforming vanilla LSTM in certain applications.[LSTM7] In 2017, Google started using similar "neural architecture search."[NAS]

Since 2006, we have worked with the software industry (e.g., LifeWare) to greatly improve handwriting recognition. In 2009, through the efforts of Alex, LSTM trained by CTC became the first RNN to win international competitions, namely, three ICDAR 2009 Connected Handwriting Competitions (French, Farsi, Arabic). This attracted enormous interest from industry. LSTM was soon used for everything that involves sequential data such as language and speech[LSTM10-11][LSTM4][DL1] and videos. By 2017, LSTM powered Facebook's machine translation (over 30 billion translations per week),[FB17][DL4] Apple's Quicktype on roughly 1 billion iPhones,[DL4] the voice of Amazon's Alexa,[DL4] Google's speech recognition (on Android smartphones since 2015)[GSR15][DL4] & image caption generation[DL4] & machine translation[GT16][DL4] & automatic email answering[DL4] etc. Business Week called LSTM "arguably the most commercial AI achievement."[AV1] The first Large Language Models (LLMs) were based on LSTM as well.

By 2016, more than a quarter of the awesome computational power for inference in Google's datacenters was used for LSTM (and 5% for another popular Deep Learning technique called CNNs—see Sec. 19).[JOU17] Google's on-device speech recognition of 2019 (now on the phone, not on the server) was still based on LSTM.

**LSTM for Healthcare.**[DEC] A simple Google Scholar search turns up innumerable medical articles that have "LSTM" in their title, e.g., for learning to diagnose, ECG time signal analysis and classification, patient subtyping, clinical concept extraction, diagnosis of arrhythmia, drug-drug interaction extraction, hospitalization prediction, monitoring on personal wearable devices, automatic pain level classification, cardiovascular disease risk factors prediction, 4D medical image segmentation, detection of radiological abnormalities, automated sleep stage classification, blood glucose prediction, diabetes detection, lung cancer detection, respiration prediction, real-time tumor tracking, breast cancer detection from histopathological images, air pollution forecasting, protein model quality assessment, protein secondary structure prediction, modeling genome data, generation of drug-like chemical matter, pandemic forecasting, Covid-19 detection, Covid-19 classification, Covid-19 prediction, and many more.

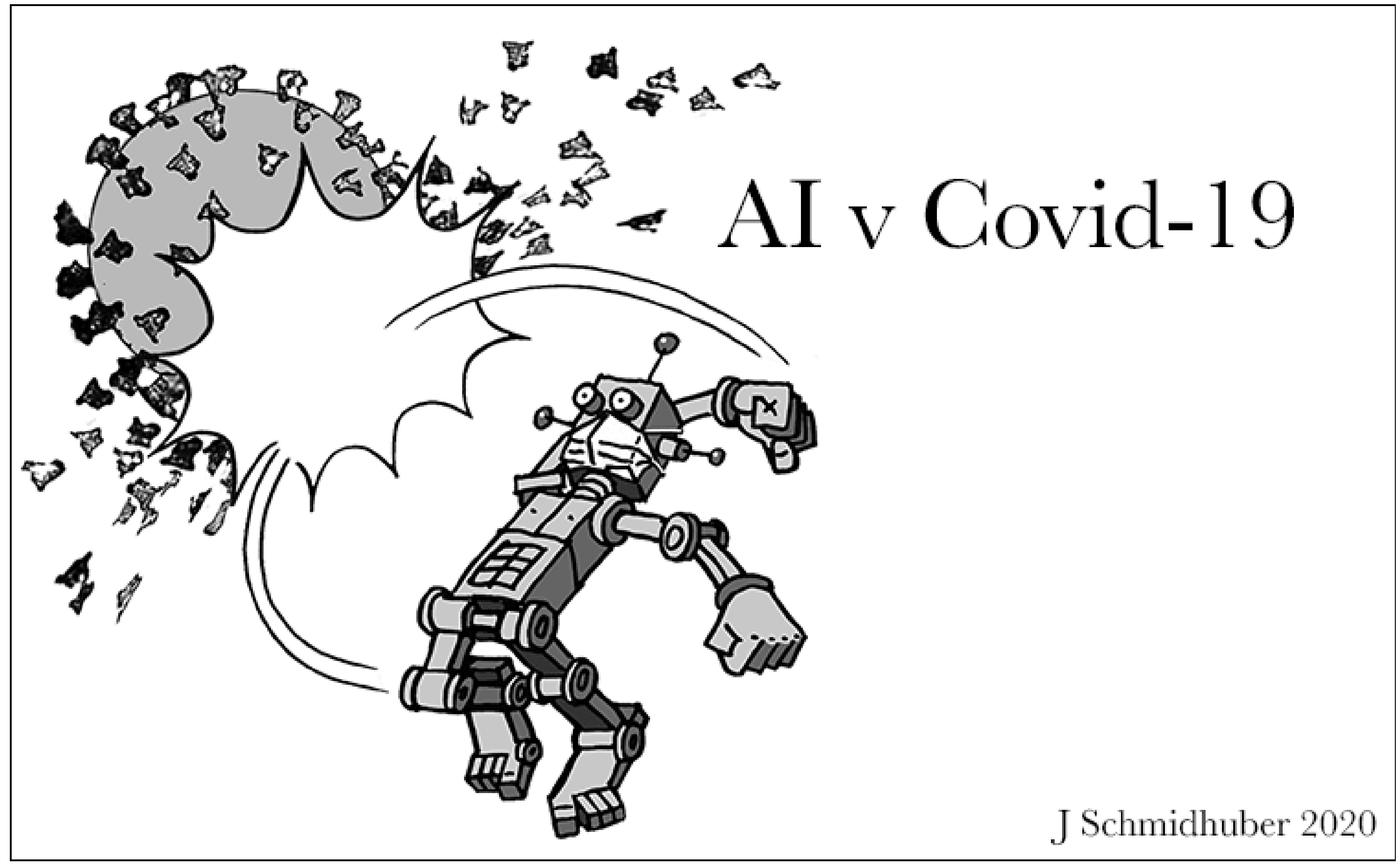

Through the work of my students Rupesh Kumar Srivastava and Klaus Greff, the LSTM principle also led to our Highway Networks[HW1] of May 2015, the first working very deep FNNs with hundreds of layers (previous NNs had at most a few tens of layers). 7 months later,

Microsoft won the ImageNet 2015 contest with an open-gated Highway Net variant called ResNet[HW2] (ResNets are like Highway Nets whose gates are always open). The earlier Highway Nets perform roughly as well as ResNets on ImageNet.[HW3] Variants of Highway gates are also used for certain algorithmic tasks, where the simpler residual layers do not work as well.[NDR] See also: who invented deep residual learning?[WHO11]

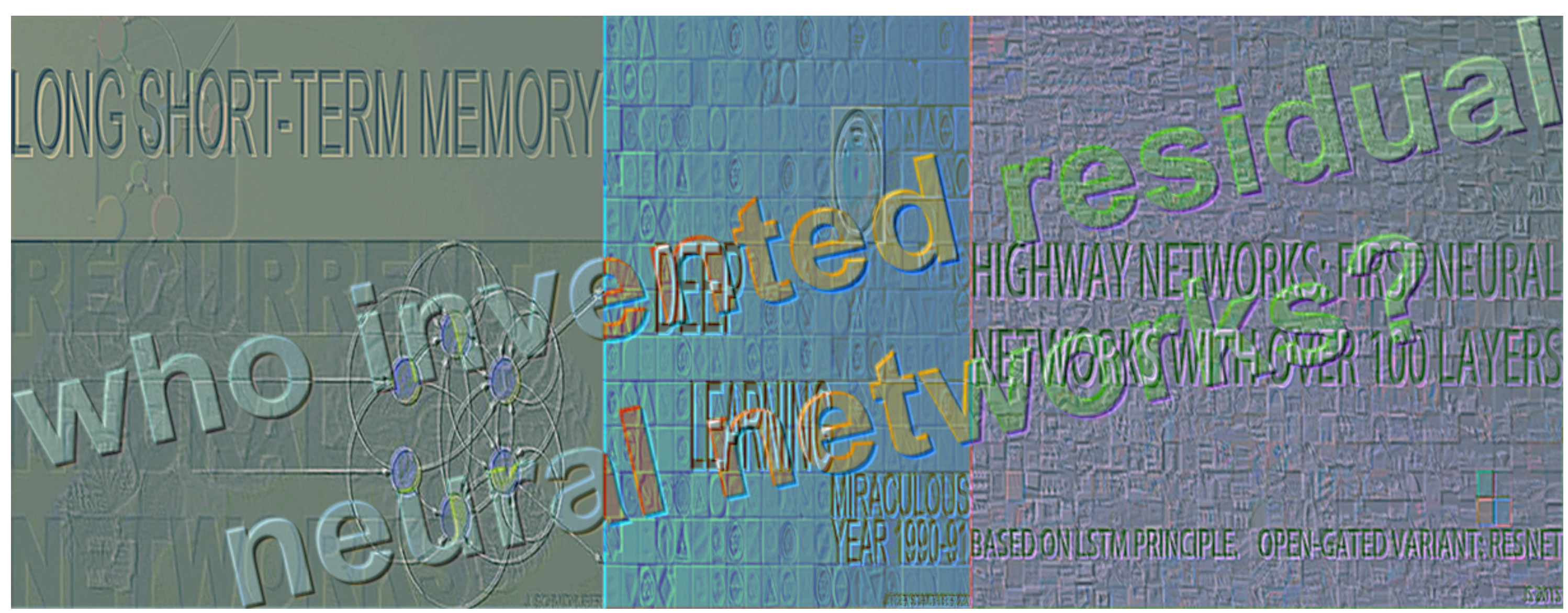

Deep learning is all about NN depth.[DL1][DLH] LSTMs brought essentially *unlimited* depth to supervised *recurrent* NNs; Highway Nets brought it to *feedforward* NNs. Interestingly, LSTM has become the most cited NN of the 20th century; an open-gated Highway Net variant the most cited NN of the 21st.[MOST][MOST26]

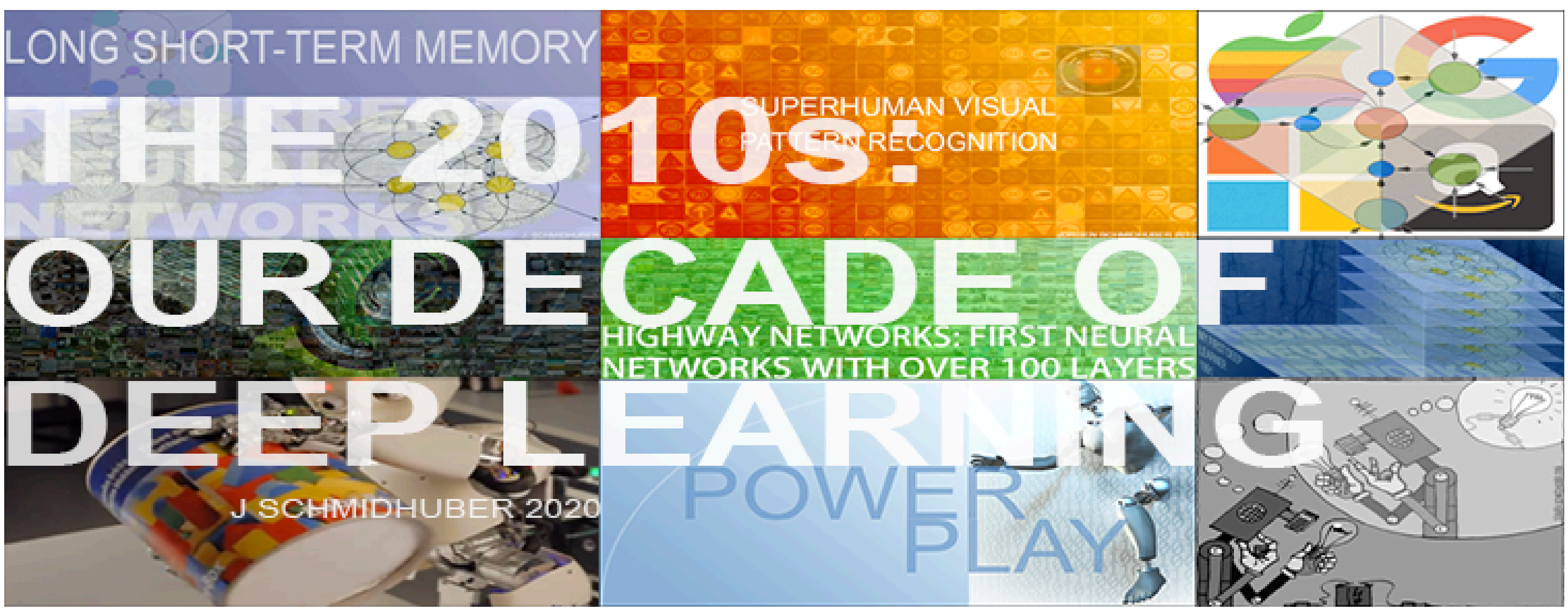

We also trained LSTM through *Reinforcement Learning* (RL) for robotics without a teacher, e.g., with my postdoc Bram Bakker[LSTM-RL] (2002). And also through Neuroevolution and policy gradients, e.g., with my PhD student Daan Wierstra,[LSTM12][RPG07][RPG][LSTMPG] who later became employee number 1 of DeepMind, the company co-founded by his friend Shane Legg, another PhD student from my lab (Shane and Daan were the first persons at DeepMind with AI publications and PhDs in computer science). RL with LSTM has become important. For example, in 2019, DeepMind beat a pro player in the game of Starcraft, which is harder than Chess or Go[DM2] in many ways, using Alphastar whose brain has a deep LSTM core trained by RL.[DM3] An RL LSTM (with 84% of the model's total parameter count) also was the core of the famous OpenAI Five which learned to defeat human experts in the Dota 2 video game (2018).[OAI2] Bill Gates called this a "huge milestone in advancing artificial intelligence."[OAI2a]

Essential foundations for all of this were laid in 1991. My team subsequently developed LSTM & CTC etc. with the help of basic funding from TU Munich and the (back then private) Swiss Dalle Molle Institute for AI (IDSIA), as well as public funding which I acquired from Switzerland, Germany, and EU during the *"Neural Network Winter"* of the 1990s and early 2000s, trying to keep the field alive when few were interested in NNs. I am especially thankful to Professors Kurt Bauknecht, Leslie Kaelbling, Ron Williams, and Ray Solomonoff whose positive reviews of my grant proposals have greatly helped to obtain financial support from SNF since the 1990s.

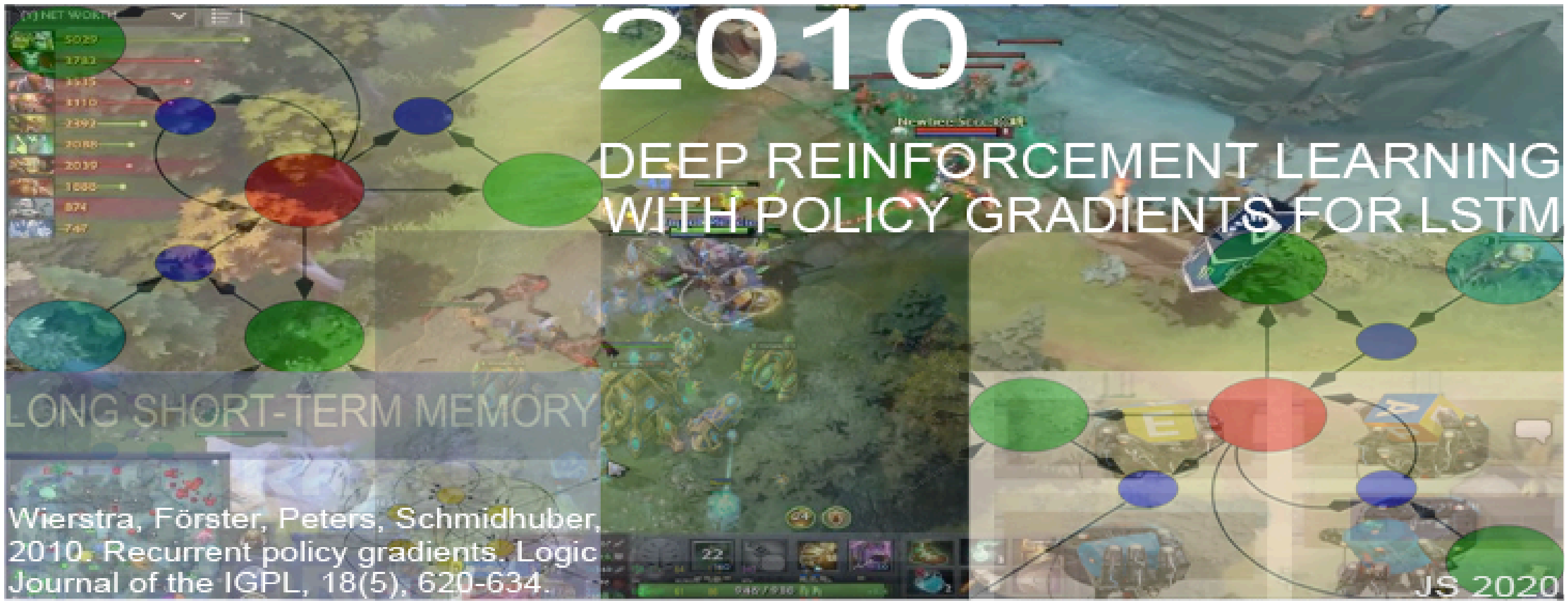

# 5. Artificial Curiosity / Generative Adversarial NNs (1990)

As humans interact with the world, they learn to predict the consequences of their actions. They are also curious, designing experiments that lead to novel data from which they can learn more. To build curious artificial agents,[AC] I introduced a new type of *active* unsupervised or self-supervised learning in 1990.[AC90,AC90b] It is based on a minimax game where one NN minimizes the objective function maximized by another NN. Today, I refer to this duel between two unsupervised adversarial NNs as Adversarial Artificial Curiosity,[AC20] to distinguish it from our later types of Artificial Curiosity since 1991 (see Sec. 6).

How does Adversarial Curiosity work? The first NN is called the controller C. C (probabilistically) generates outputs that may influence an environment. The second NN is called the world model M. It predicts the environmental reactions to C's outputs. Using gradient descent, M minimizes its error, thus becoming a better predictor. But in a zero sum game, C tries to find outputs that maximize the error of M. M's loss is the gain of C.

That is, C is motivated to invent novel outputs or experiments that yield data that M still finds surprising, until the data becomes familiar and eventually boring. Compare more recent summaries and extensions of this principle.[AC09] A 2010 survey[AC10] summarised the work of 1990 as follows: a *"neural network as a predictive world model is used to maximize the controller's intrinsic reward, which is proportional to the model's prediction errors."*

So in 1990 we already had unsupervised or self-supervised neural nets that were both *generative* and *adversarial* (using much later terminology from 2014[GAN14]), generating experimental outputs yielding novel data, not only for stationary patterns but also for pattern sequences, and even for the general case of Reinforcement Learning (RL).

The popular *Generative Adversarial Networks (GANs)*[GAN90-25] (2010-2014) are an Instance of Adversarial Curiosity[GAN90] where the environment simply returns whether C's current output is in a given set.[GAN20-25][DLP]

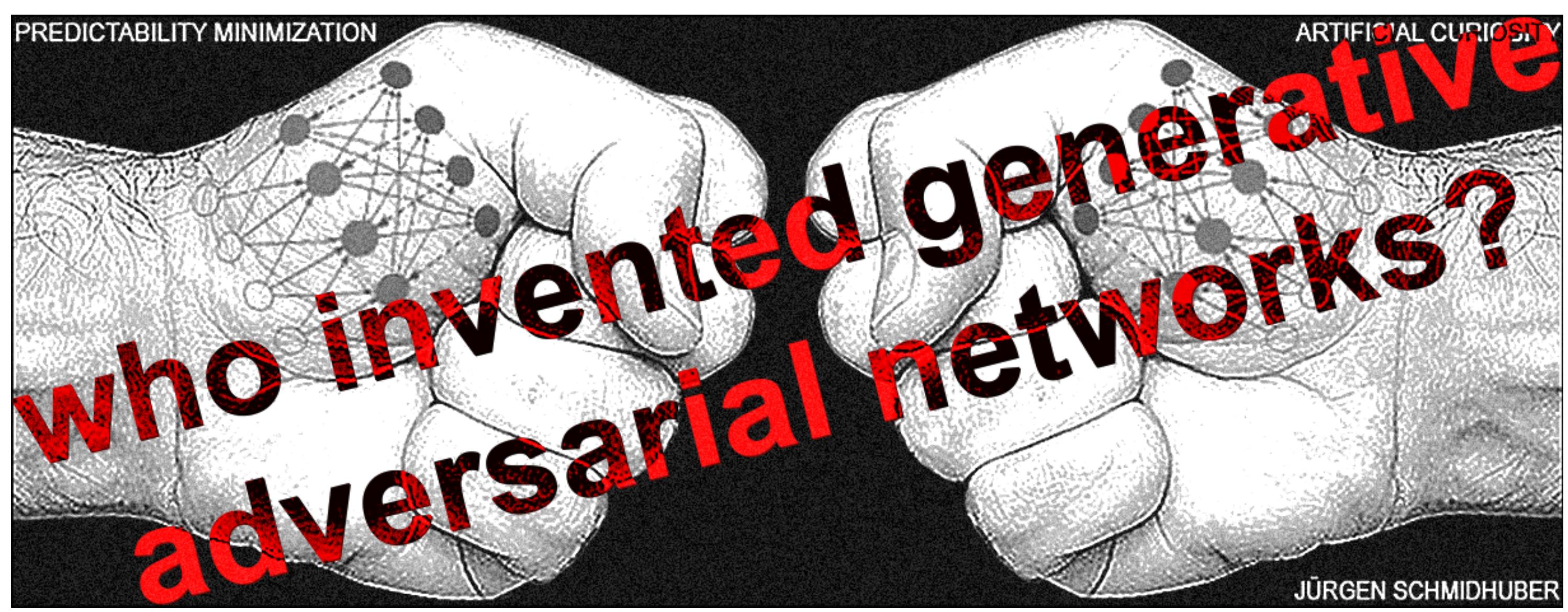

BTW, note that the closely related Adversarial Curiosity[GAN90,90b] & GANs[GAN14] & Adversarial *Predictability Minimization* (see Sec. 7) are very different from other early adversarial machine learning settings[S59][H90] which neither involved unsupervised NNs nor were about modeling data nor used gradient descent.[GAN20] See also: who invented generative adversarial networks?[WHO8]

# 6. Artificial Curiosity Through NNs That Maximize Learning Progress (1991)

Numerous improvements of the original Adversarial Curiosity of 1990 (AC1990, Sec. 5) are summarized in more recent surveys.[AC06][AC09][AC10][AC] Here I focus on the first important improvement of 1991.[AC91][AC91b]

The errors of AC1990's world model M (to be minimized, Sec. 5) are the rewards of the controller C (to be maximized). This makes for a fine exploration strategy in many deterministic environments. In stochastic environments, however, this might fail. C might learn to focus on situations where M can always get high prediction errors due to randomness, or due to its computational limitations. For example, an agent controlled by C might get stuck in front of a TV screen showing highly unpredictable white noise.[AC10]

Therefore, as pointed out in 1991, in stochastic environments, C's reward should not be the errors of M, but (an approximation of) the *first derivative* of M's errors across subsequent training iterations, that is, M's *improvements*.[AC91][AC91b] As a consequence, despite its high errors in front of the noisy TV screen above, C won't get rewarded for getting stuck there. Both the totally predictable and the fundamentally unpredictable will get boring. This insight led to lots of follow-up work[AC10] on artificial scientists and artists.[AC09][AC]

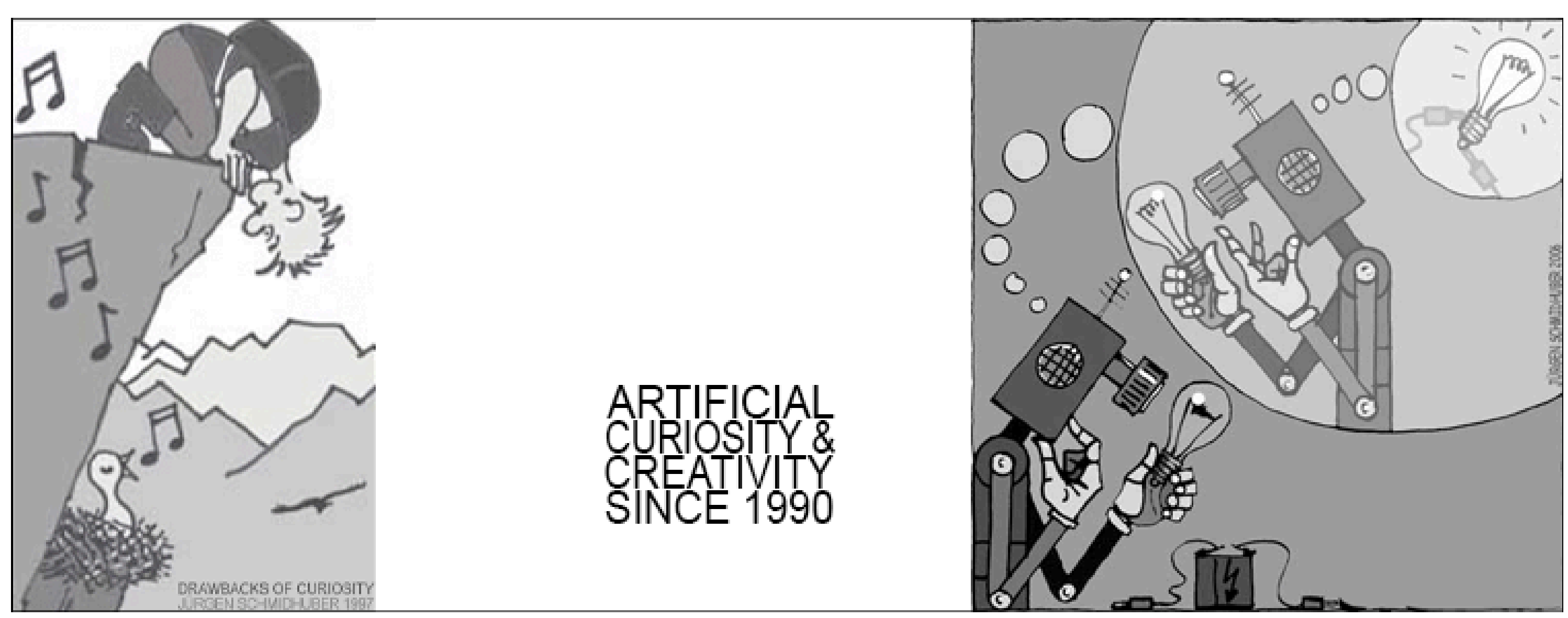

## 7. Adversarial Networks Create Disentangled Data Representations (1991)

Soon after my first work on adversarial generative networks in 1990 (see Sec. 5), I introduced a variation of the unsupervised adversarial minimax principle while I was a postdoc at the University of Colorado at Boulder. One of the most important NN tasks is to learn the statistics of given data such as images. To achieve this, I used again the principles of gradient descent/ascent in a minimax game where one NN minimizes the objective function maximized by another. This duel between two unsupervised adversarial NNs was called Predictability Minimization (PM, 1990s).[PM0-2] Contrary to later claims,[GAN14] PM is indeed a pure minimax game, e.g., Equation 2 of [PM2].[GAN20][T20][DLP]

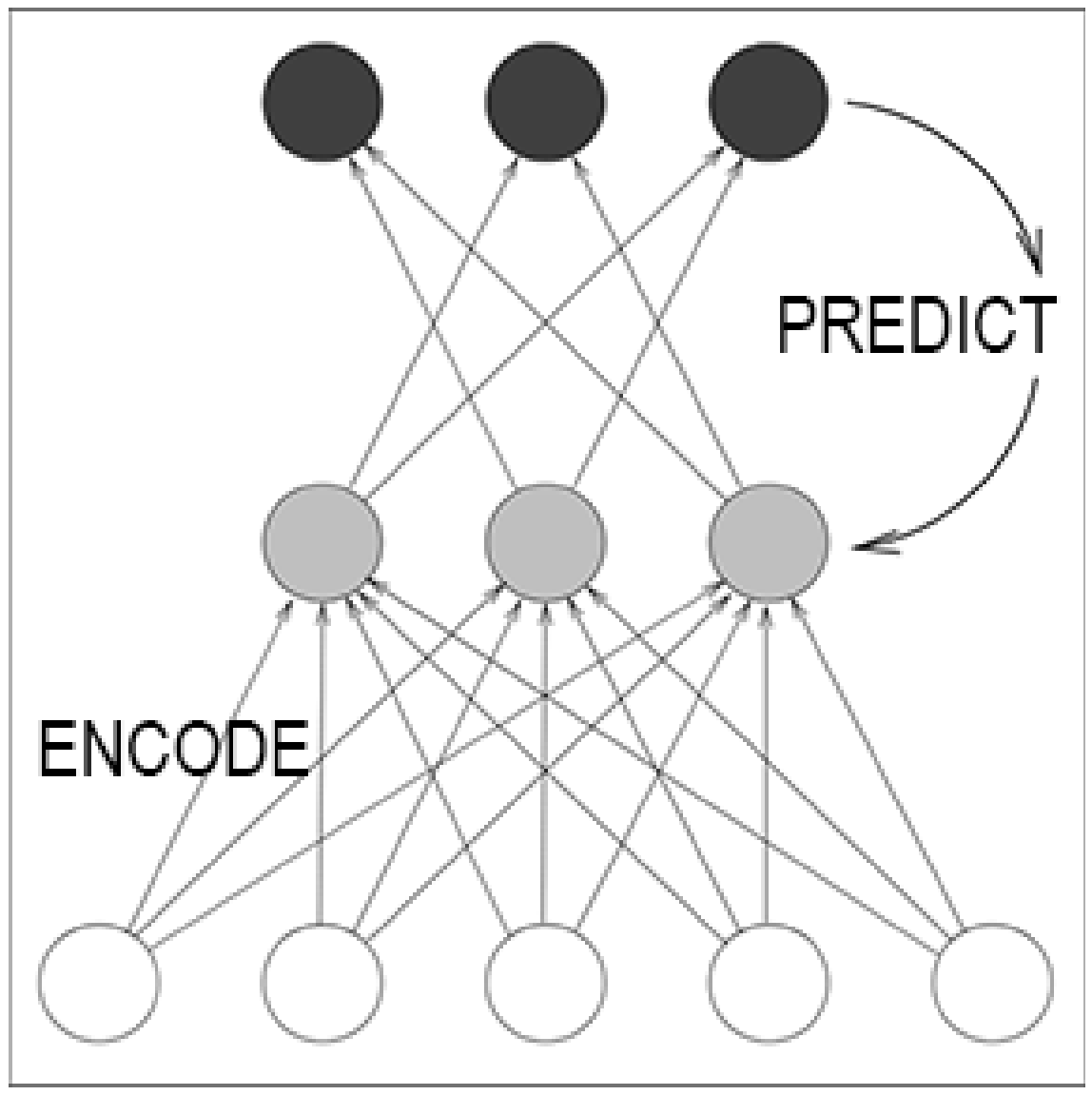

The first toy experiments with PM[PM1] were conducted nearly three decades ago when compute was about a million times more expensive than today. When it had become about 10 times cheaper 5 years later, we could show that semi-linear PM variants applied to images automatically generate feature detectors well-known from neuroscience, such as on-center-off-surround detectors, off-center-on-surround detectors, and orientation-sensitive bar detectors.[PM2]

# 8. Unnormalized Linear Transformers (1991). Fast Weight Programmers: NNs Learn to Program NNs

The first Large Language Models (LLMs) were based on LSTM (see Sec. 4). However, in the late 2010s, despite their limited time windows, recurrent-connection-free **Transformers**[TR1,TR25] (see the T in ChatGPT[GPT3]) started to excel at Natural Language Processing (NLP), a traditional LSTM domain. Remarkably, Transformers also have their roots in our Miraculous Year of 1991!

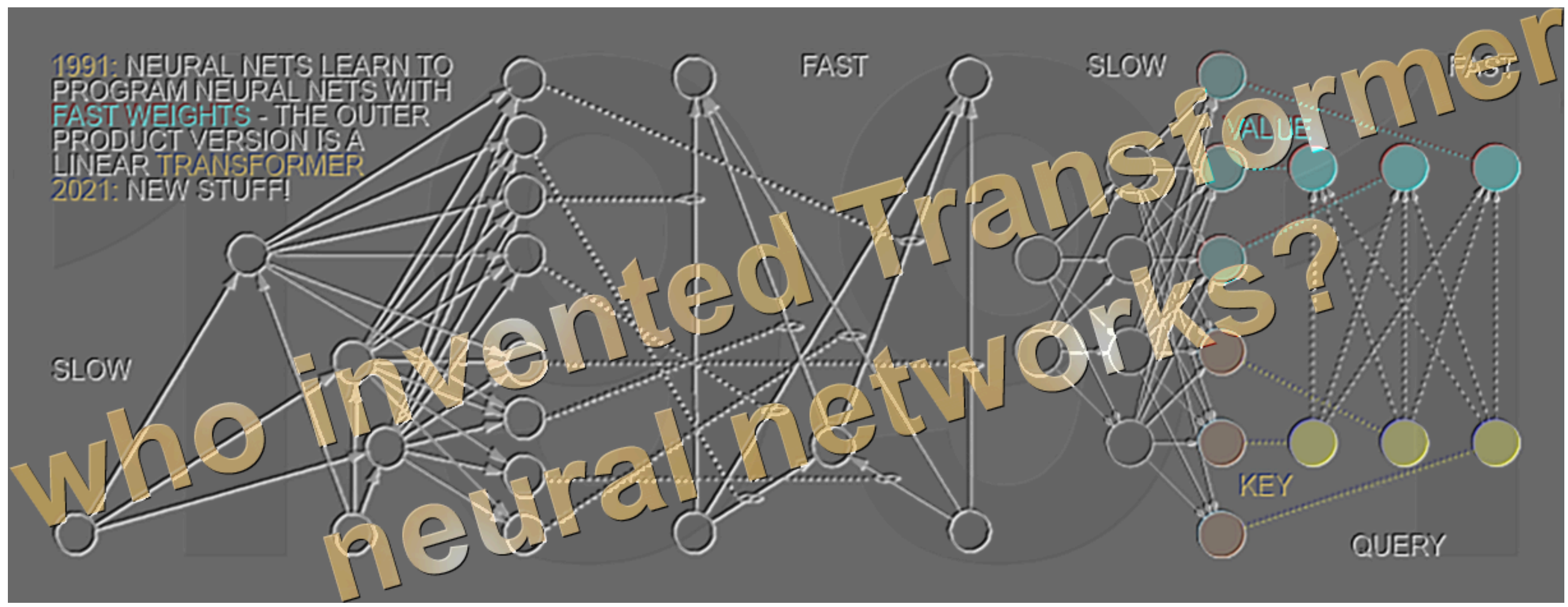

In March 1991, when compute was a million times more expensive than in 2022, even before the LSTM, I published the first Transformer variant, which is now called the unnormalized linear Transformer (ULTRA).[ULTRA][FWP0] It had to be more efficient than Google's 2017 *quadratic* Transformer:[TR1][TR25] ULTRA's computational costs scale *linearly* in input size, rather than *quadratically* (in 1991, no journal would have accepted an NN that scales quadratically). My 1993 paper on recurrent ULTRA extensions[FWP2] talked about learning *"internal spotlights of attention"*—compare the recent attention terminology, e.g., *"attention is all you need,"*[TR1] and tweets of 2022 & 2023.

The ULTRA was a by-product of more general research on NNs learning to program other NNs. A typical NN has many more connections than neurons. In traditional NNs, neuron activations change quickly, while connection weights change slowly. That is, the numerous weights cannot implement short-term memories or temporal variables, only the few neuron activations can. Non-traditional NNs with quickly changing *"fast weights"* overcome this limitation. Dynamic links or fast weights for NNs were introduced by Christoph v. d. Malsburg in 1981[FAST] and further studied by others.[FASTa,b] However, before 1991, no network learned by gradient descent to quickly compute the changes of the fast weight storage of another network or of itself.

Enter the end-to-end-differentiable Fast Weight Programmers (FWPs)[FWP] published in 1991-93.[FWP0-2] There a slow NN learns to program the weights of a separate fast NN. That is, I *separated storage and control* like in traditional computers, but in a fully neural way (rather than in a hybrid fashion[PDA1-2][DNC]). FWPs embody the principles found in what is now called

attention[ATT] and *Transformers*.[TR1-6][FWP] In fact, one of my FWPs is the 1991 unnormalized linear Transformer above (with what's now called "linearized self-attention").[TR5-6][FWP][ATT] See this tweet celebrating the 30-year anniversary of the 1992 journal publication.[FWP0-1]

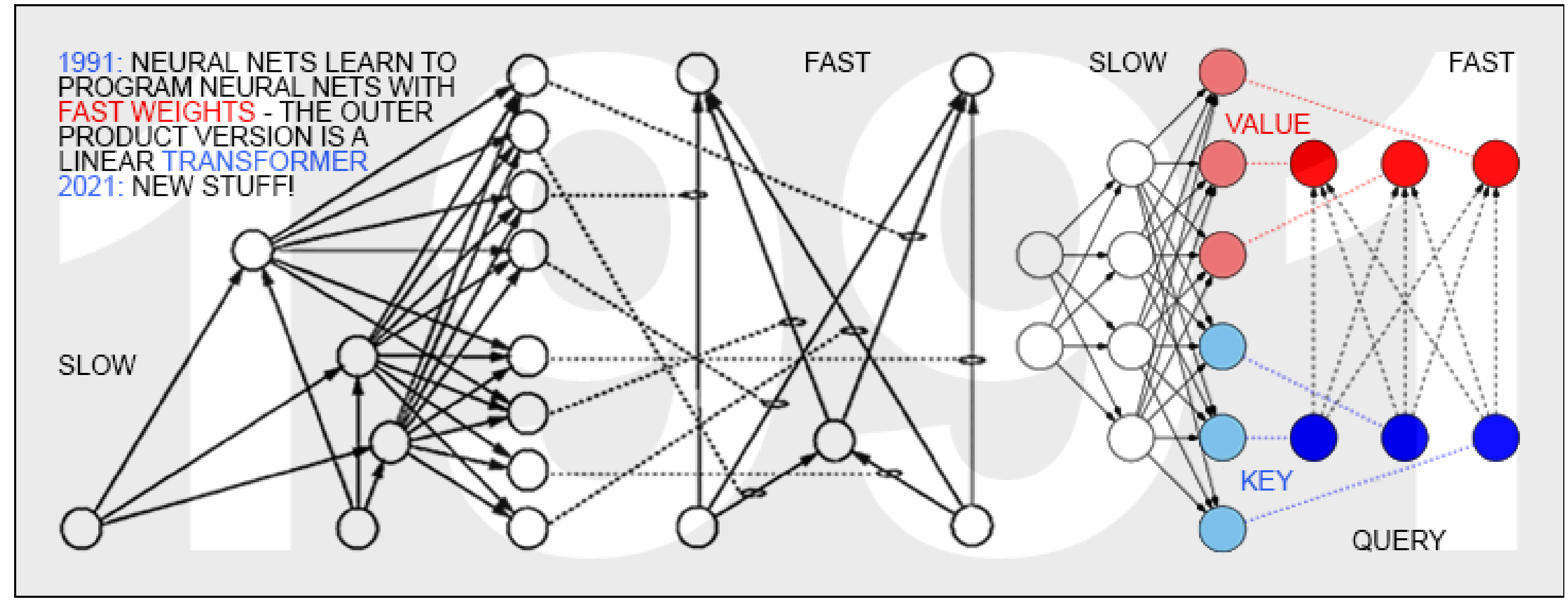

Some of my FWPs used gradient descent-based, active control of fast weights through 2D tensors or outer product updates[FWP1-2] (compare our more recent work on this[FWP3-3a][FWP6-7]). One of the motivations[FWP2] was to get many more temporal variables under end-to-end differentiable control than what's possible in standard RNNs of the same size: O(H^2) instead of O(H), where H is the number of hidden units. A quarter century later, others followed this approach.[FWP4a][DLP] The 1993 paper[FWP2] also explicitly introduced the "modern" terminology of learning internal spotlights of attention in end-to-end-differentiable networks. Compare Sec. 9 on learning attention.

Work on extensions of ULTRAs and other FWPs has become mainstream research, aiming to develop sequence models that are both efficient and powerful.[TR6,TR6a][LT23-25][FWP23-25b] See also: who invented transformer neural networks?[WHO10]

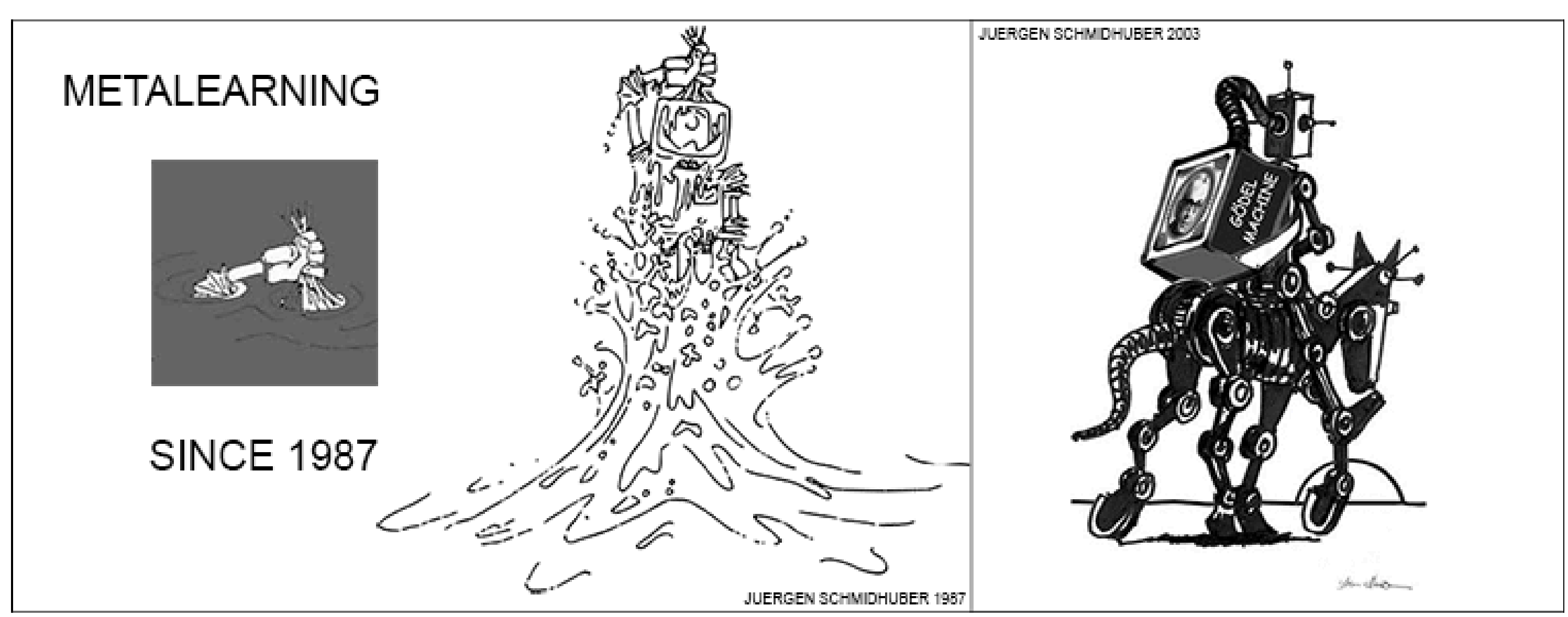

I also showed how FWPs can be used for meta-learning or learning to learn, one of my main research topics since 1987.[META1][META] In follow-up work[FWPMETA1-10] since 1992, the slow RNN and the fast RNN are *identical:* the initial weight of each connection in the net is trained by gradient descent, but during an episode, each connection can be addressed and read and modified by the net itself (through *O(log n)* special output units where *n* is the number of connections), and the connection's weight may rapidly change—the network becomes *self-referential* in the sense that it can in principle learn to run arbitrary computable weight change algorithms or learning algorithms (for all of its weights) *on itself*. This led to many follow-up papers in the 1990s and 2000s.

Deep Reinforcement Learning (RL) without a teacher can also profit from fast weights even when the system's dynamics are not differentiable, as shown in 2005 by my former postdoc Faustino Gomez[FWP5] (now CEO of NNAISENSE) when affordable computers were about 1000 times faster than in the early 1990s.

2005: 1st paper with "learn deep" in the title

(on deep reinforcement learning with recurrent nets & neuroevolution)

Interestingly, our related work on Deep RL in the same year (but without fast weights) to my knowledge was the first machine learning publication with the word combination "learn deep" in the title[DL6-6a] (2005; soon afterwards many started talking about "Deep Learning").

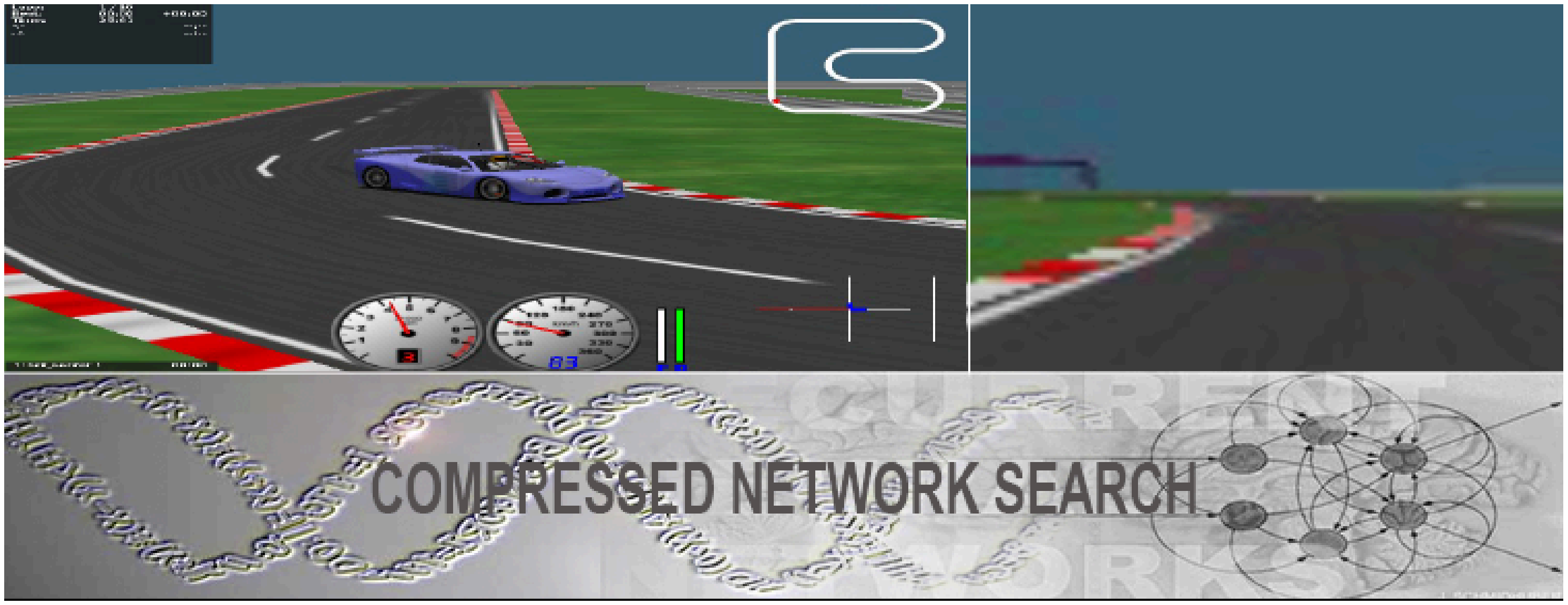

Over the decades we have published quite a few additional ways of learning to generate quickly numerous weights of large NNs through very compact codes.[KO0-2][CO1-3] Here we exploited that the Kolmogorov complexity or algorithmic information content of successful huge NNs may actually be rather small. In particular, in July 2013, Compressed Network Search[CO2] was the first Deep Learning model to successfully learn control policies directly from high-dimensional sensory input (video) using reinforcement learning, without any unsupervised pre-training (unlike in Sec. 1). Soon afterwards, DeepMind also had a Deep RL system for high-dimensional sensory input.[DM1-2]

# 9. Learning Sequential Attention with NNs (1990)

Unlike traditional NNs, humans use sequential gaze shifts and selective attention to detect and recognize patterns. This can be much more efficient than the highly parallel approach of traditional FNNs. That's why we introduced sequential attention-learning NNs three decades ago (1990 and onwards).[ATT0-1] Shortly afterwards, I also explicitly addressed the learning of *"internal spotlights of attention"* in RNNs[FWP2] (see Sec. 8).

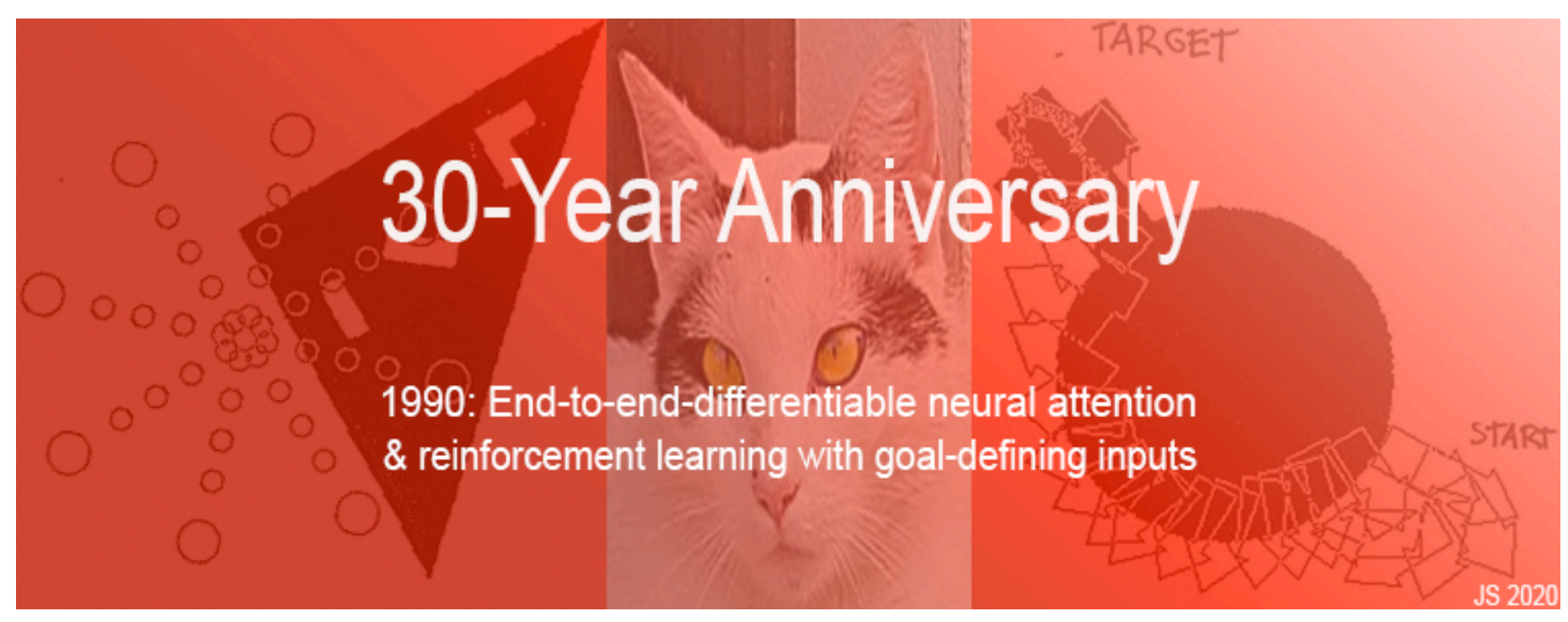

So back then we already had *both* of the now common types of neural sequential attention: end-to-end-differentiable *"soft"* attention (in *latent* space) through multiplicative units within NNs,[FWP][FWP2] and *"hard"* attention (in *observation* space) in the context of Reinforcement Learning (RL).[ATT0-1] In particular, in 1991, I had what's now called the unnormalised linear Transformer[ULTRA] with linearized self-attention.[TR1-6][FWP0-1][FWP6][FWP] (see Sec. 8). This led to lots of follow-up work. Today, many are using sequential attention-learning NNs.

My overview paper for CMSS 1990[ATT2] summarised in Section 5 our early work on attention, to my knowledge the first implemented neural system for combining glimpses that jointly trains a recognition & prediction component with an attentional component (the fixation controller). As with the work of Sepp (see Sec. 3), this has often been misattributed. In 2010, the reviewer of my 1990 paper wrote about his own work:[ATT3] *"To our knowledge, this is the first implemented system for combining glimpses that jointly trains a recognition component ... with an attentional component (the fixation controller)"*[DLP] (see also Sec. 10).

# 10. Gradient Descent Finds Subgoals for Hierarchical Reinforcement Learning (1990)

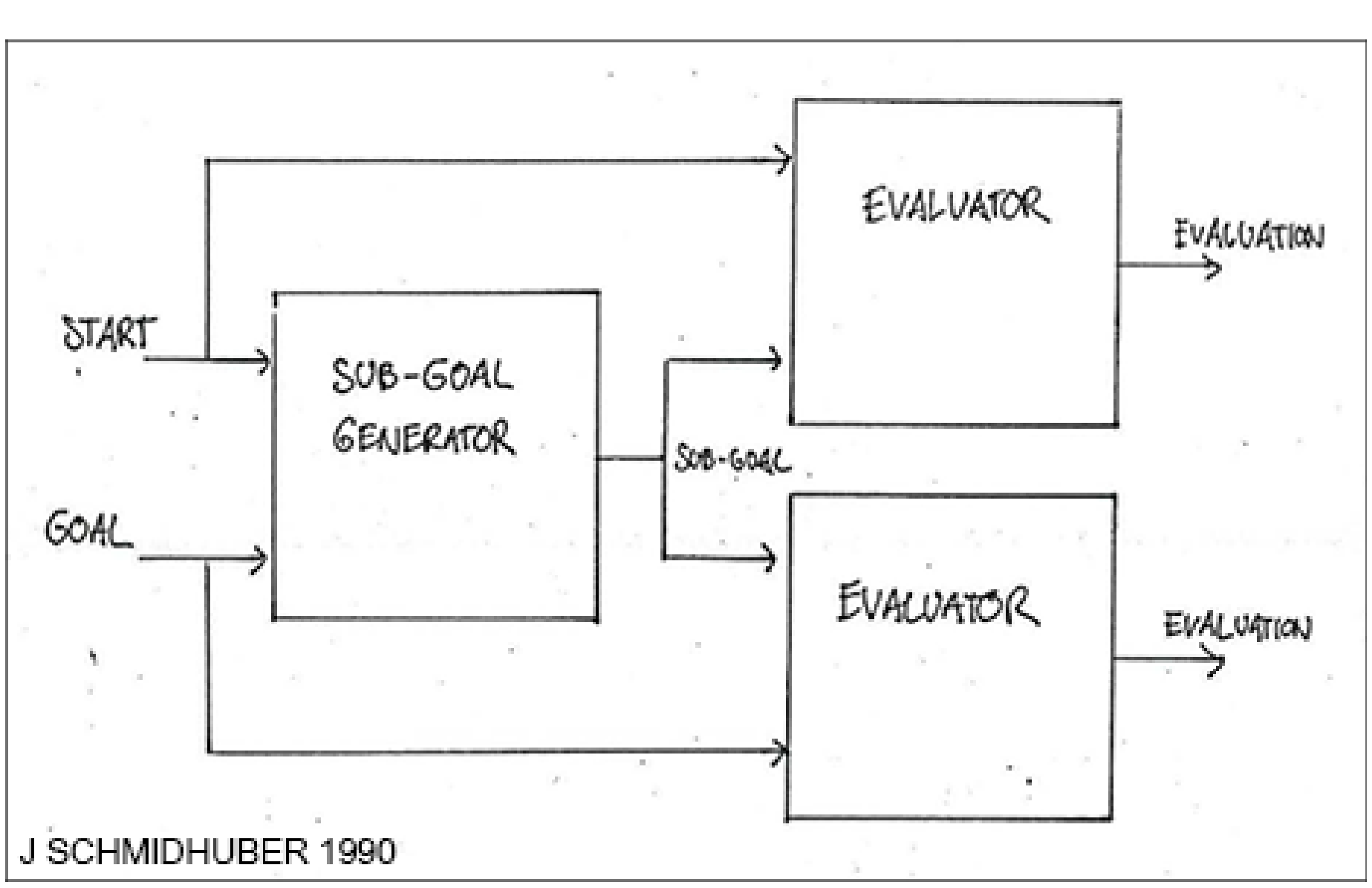

Traditional Reinforcement Learning (RL) without a teacher does not hierarchically decompose problems into easier subproblems.[HRLW] That's why in 1990 I introduced

*Hierarchical RL* (HRL) with end-to-end differentiable NN-based subgoal generators,[HRL0] also with recurrent NNs that learn to generate sequences of subgoals.[HRL1-2][LEC]

An RL machine gets extra inputs of the form *(start, goal)*. An evaluator NN learns to predict the rewards/costs of going from *start* to *goal*. An (R)NN-based subgoal generator also sees *(start, goal)*, and uses (copies of) the evaluator NN to learn by gradient descent a sequence of cost-minimising intermediate subgoals. The RL machine tries to use such subgoal sequences to achieve final goals. The system is learning action plans at multiple levels of abstraction and multiple time scales. At least in principle, it solves what was called an "open problem" in 2022.[LEC]

Our 1990-91 papers[HRL0-1] started a series of papers on HRL.(e.g., [HRL4]) The reviewer of my 1990 paper[ATT2] (which summarised in Section 6 our early work on HRL) was last author of a 1992 paper on HRL.[HRL3] Compare Sec. 9.

# 11. Planning with Recurrent World Models (1990)

The concept of a *mental model of the world*—a *world model*—dates back millennia. Plato suggested[PLA1] that we recognize objects by recollecting internal *blueprints* or *templates*, today often called *internal representations*. Aristotle wrote that *phantasia* or *mental images* allow humans to imagine the future and to plan action sequences by mentally manipulating images in the absence of the actual objects.[ARI1] Only 2,370 years later—a mere blink of an eye by cosmical standards—we are witnessing a boom in world models based on NNs for AI in the physical world.[WM26][WM26b] New startups on this are emerging.

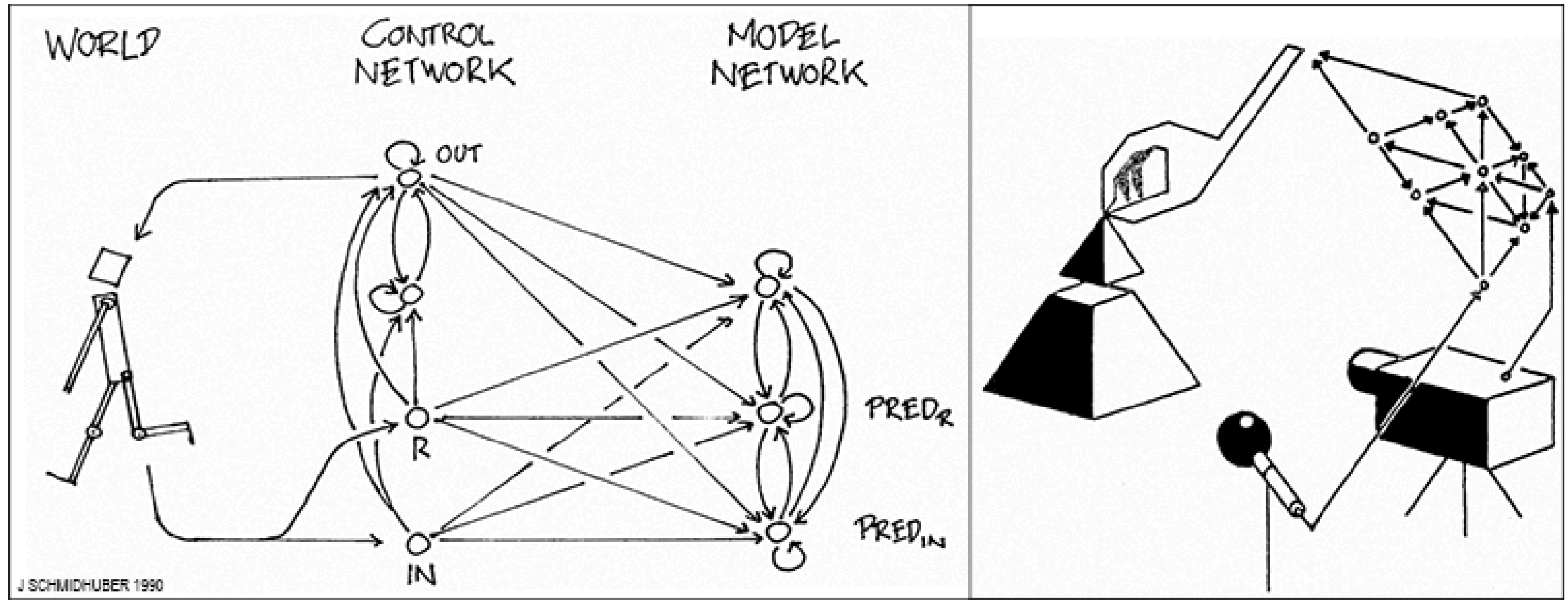

Much of this goes back to 1990, when I introduced RNNs as general purpose world models.[AC90][PLAN2][PLAN3] I studied adaptive agents living in *partially observable environments* where non-trivial kinds of *memory* are required to act successfully. I used the term *world model* for an RNN that learns to predict the agent's sensory inputs (including pain and reward signals) reflecting the consequences of the actions of a separate *controller* RNN steering the agent. The controller C used the world model M to plan its action sequences through *mental experiments*.[PLAN] Since RNNs are general purpose computers, this approach went beyond previous, less powerful, FNN-based systems (since 1987) for *fully* observable environments.[WER87-89][MUN87][NGU89]

In the beginning, my 1990 world model M knew nothing. That's why my 1990 controller C (a generative model with stochastic neurons) was *intrinsically motivated* through adversarial artificial curiosity[AC90][AC90b][AC] to invent action sequences or *experiments* that yield data from which M can learn something: C simply tried to *maximize* the prediction error *minimized* by M. Today, they call this a generative adversarial network (GAN).[GAN90-25] See also Sec. 5 and Sec. 6.

The 1990 system didn't learn like today's *foundation models* and *large language models* (LLMs) by downloading and imitating the web. No, it generated *its own self-invented experiments* to collect *limited but relevant* data from the environment, like a physicist, or a baby.[AC90,AC90b][AC] It was a simple kind of *artificial scientist*.

This led to lots of follow-up publications, also in recent years.[PLAN-PLAN6][LEC][WM26] In 2014, we even founded an AGI company for *Physical AI* in the real world, based on neural world models.[NAI] It achieved lots of remarkable milestones in collaboration with world-famous companies. Alas, like some of our projects, the company may have been a bit ahead of time, because real world robots and hardware are so challenging. Nevertheless, it's great that in the 2020s, new world model startups have been created!

The 1990 FKI report[AC90] also introduced several other concepts that have become popular. See Sec. 12, Sec. 13, Sec. 14, Sec. 5, Sec. 20.

# 12. Goal-Defining Commands as Extra NN Inputs (1990)

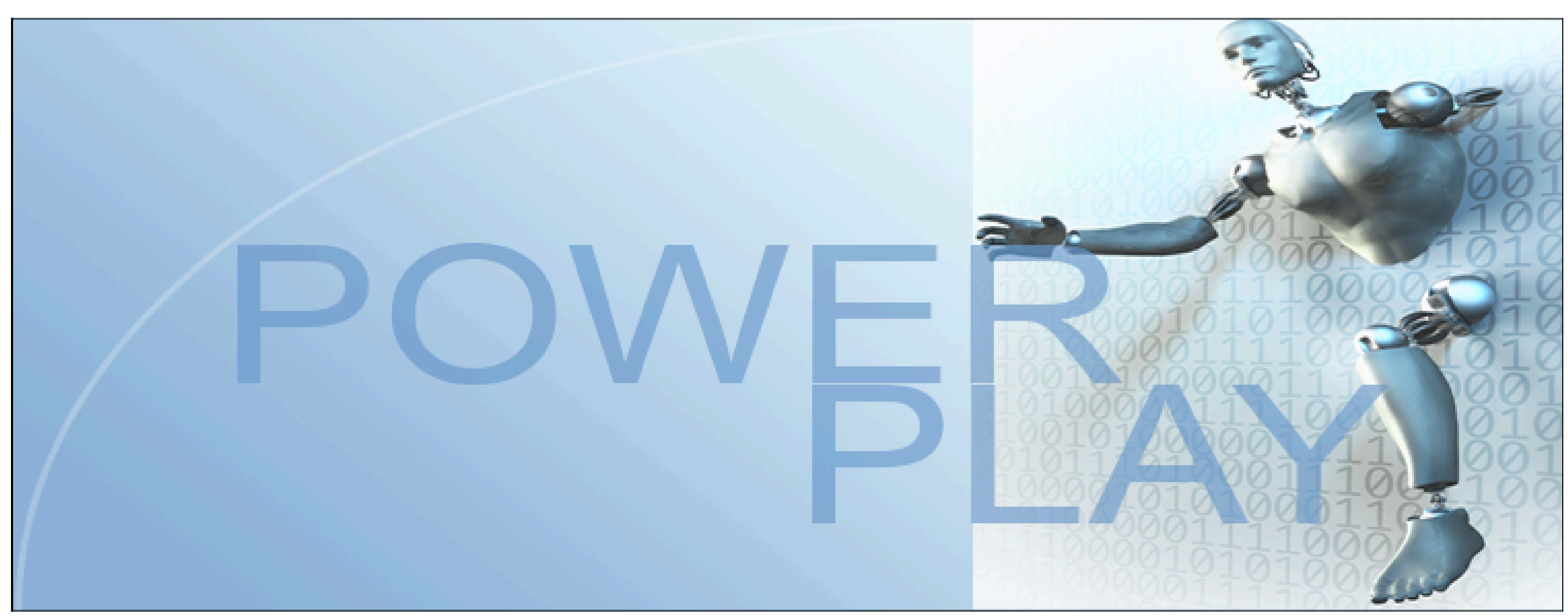

One concept that is widely used in today's RL NNs are extra *goal-defining input patterns* that encode various tasks, such that the NN knows which task to execute next. We introduced this in 1990 in various contexts.[ATT0-1][HRL0-1] In [ATT0-1] a reinforcement learning neural controller learned to control a fovea through sequences of saccades to find particular objects in visual scenes, thus learning sequential attention (see Sec. 9). User-defined goals were provided to the system by special "goal input vectors" that remained constant (Sec. 3.2 of [ATT1]) while the system shaped its stream of visual inputs through fovea-shifting actions.

Hierarchical RL (HRL, Sec. 10) with end-to-end differentiable subgoal generators[HRL0-1][LEC] also uses an NN with task-defining inputs of the form *(start, goal)*, learning to predict the costs of going from start to goal. (Compare my former student Tom Schaul's "universal value function approximator" at DeepMind a quarter century later.[UVF15])

This led to lots of follow-up work. For example, our POWERPLAY RL system (2011)[PP][PP1] also uses task-defining inputs to distinguish between tasks, continually *inventing on its own new goals and tasks*, incrementally learning to become a more and more general problem solver in an active, partially unsupervised or self-supervised fashion. RL robots with high-dimensional video inputs and intrinsic motivation (like in PowerPlay) learned to explore in 2015.[PP2]

# 13. High-Dimensional Reward Signals As NN Inputs (1990)

Traditional RL is based on *one-dimensional* reward signals. Humans, however, have millions of informative sensors for different types of pain and pleasure etc. To my knowledge, the 1990 tech report[AC90] was the first paper on RL with *multi-dimensional, vector-valued* pain and reward signals coming in through many different sensors, where cumulative values are predicted for all those sensors, not just for a single scalar overall reward. This later emerged again in RL in what we now call *general value functions*.[GVF] Unlike previous adaptive critics, the one of 1990[AC90] was multi-dimensional and recurrent. Unlike in traditional RL, those reward signals were also used as informative *inputs* to the controller NN learning to execute actions that maximise cumulative reward.[PLAN][LEC][DLH]

Note that such a setup is essential for meta learning or learning to learn, one of my main research topics since 1987.[META1][META][FWPMETA1-10] A meta learner must see its own reward/pain/error signals as inputs, otherwise it cannot learn better learning algorithms that map such signals to useful weight changes or policy changes!

# 14. Deterministic Policy Gradients (1990)

The section *"Augmenting the Algorithm by Temporal Difference Methods"* of the 1990 paper[AC90] also combined the *Dynamic Programming*-based *Temporal Difference* method[TD] for predicting cumulative (possibly multi-dimensional, Sec. 13) rewards with a gradient-based predictive model of the world (see Sec. 11), to compute weight changes for the separate control network. See also Sec. 2.4 of the 1991 follow-up paper[PLAN3] (and compare [NAN1]). A quarter century later, a variant of this emerged as *Deterministic Policy Gradient* algorithm (DPG) at DeepMind.[DPG][DDPG]

# 15. Networks Adjust Networks / Synthetic Gradients (1990)

In 1990, I proposed various NNs that learn to adjust other NNs,[NAN1] a prelude to the 1991 end-to-end-differentiable Fast Weight Programmers of Sec. 8. Here I focus on the section *"An Approach to Local Supervised Learning in Recurrent Networks"*.[NAN1] The global error measure to be minimized is the sum of all errors received at an RNN's output units over time. In conventional *backpropagation through time* (see surveys[BPTT1-2]), each unit needs a stack for remembering past activations which are used to compute contributions to weight changes during the error propagation phase. Instead of allowing unlimited storage capacities in the form of stacks, I introduced a second adaptive NN that learns to associate states of the RNN with corresponding error vectors. These locally estimated error gradients (rather than the true gradients) are used to adjust the RNN.[NAN1-4] Unlike standard backpropagation, the method is local in space and time.[BB1-2][NAN1] A quarter century later, DeepMind called this "Synthetic Gradients."[NAN5]

## 16. $O(n^3)$ Gradients for Online Recurrent NNs (1991)

The original 1987 fixed-size storage learning algorithm for fully recurrent continually running networks[ROB] requires $O(n^4)$ computations per time step, where n is the number of non-input units. I published a method which computes exactly the same gradient and requires fixed-size storage of the same order as the previous algorithm. But, the average time complexity per time step is only $O(n^3)$.[CUB1-2] *However, this work does not really count, since the great RNN pioneer Ron Williams had derived this method first!*[CUB0]

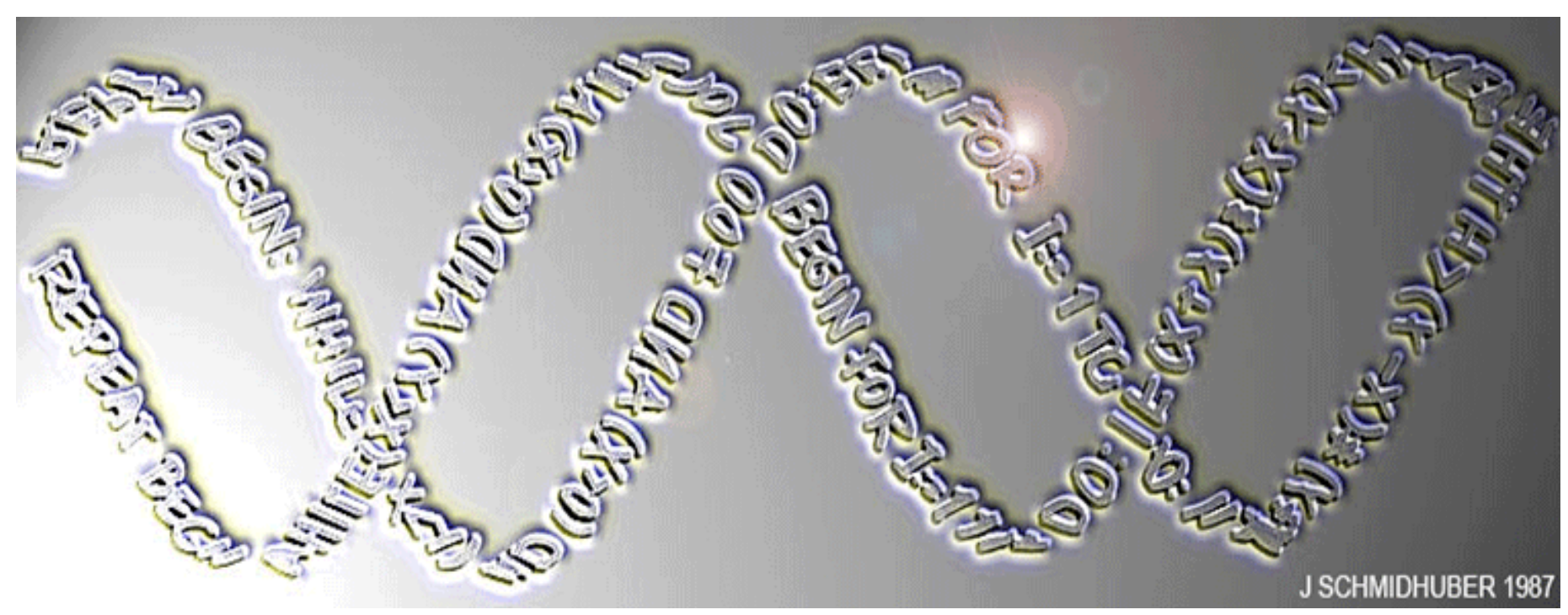

BTW, I committed a similar error in 1987 when I published what I thought was the first paper on Genetic Programming (GP), that is, on automatically evolving computer programs[GP1][GP] (authors in alphabetic order). Only later I found out that Nichael Cramer had published GP already in 1985[GP0] (and that Stephen F. Smith had proposed a related approach as part of a larger system[GPA] in 1980). Since then I have been trying to do the right thing and correctly attribute credit. At least our 1987 paper[GP1] seems to be the first on GP for codes with loops and codes of variable size, and the first on GP implemented in a Logic Programming language.

## 17. The Deep Neural Heat Exchanger (1990)

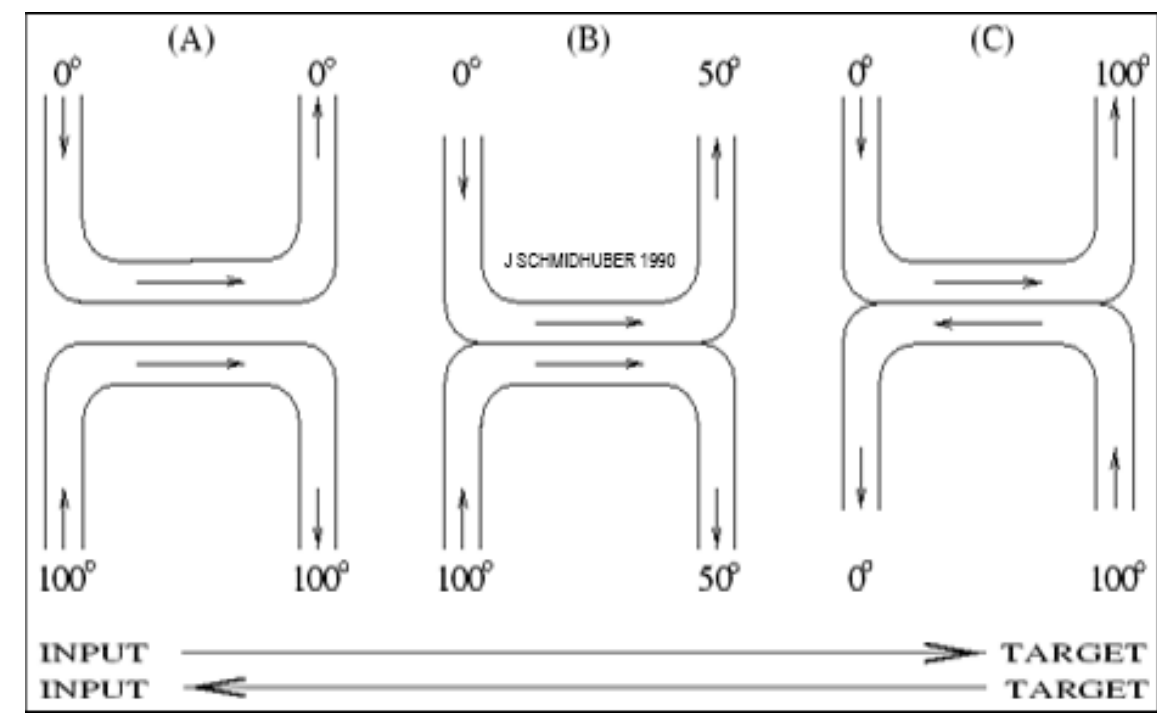

The *Neural Heat Exchanger* is supervised learning method for deep multi-layer NNs. It is inspired by the physical heat exchanger. Inputs "heat up" while being transformed through many successive layers, targets enter from the other end of the deep pipeline and "cool down." Unlike backpropagation, the method is entirely local in space and time.[BB1-2][NAN1][NOB25a] This makes its parallel implementation trivial. It was first presented during occasional talks at various universities since 1990,[NHE] and is closely related to the later Helmholtz Machine.[HEL]

[DLP] Again, experiments were conducted by my brilliant student Sepp Hochreiter (see Sec. 3, Sec. 4).

# 18. PhD Thesis (1990)

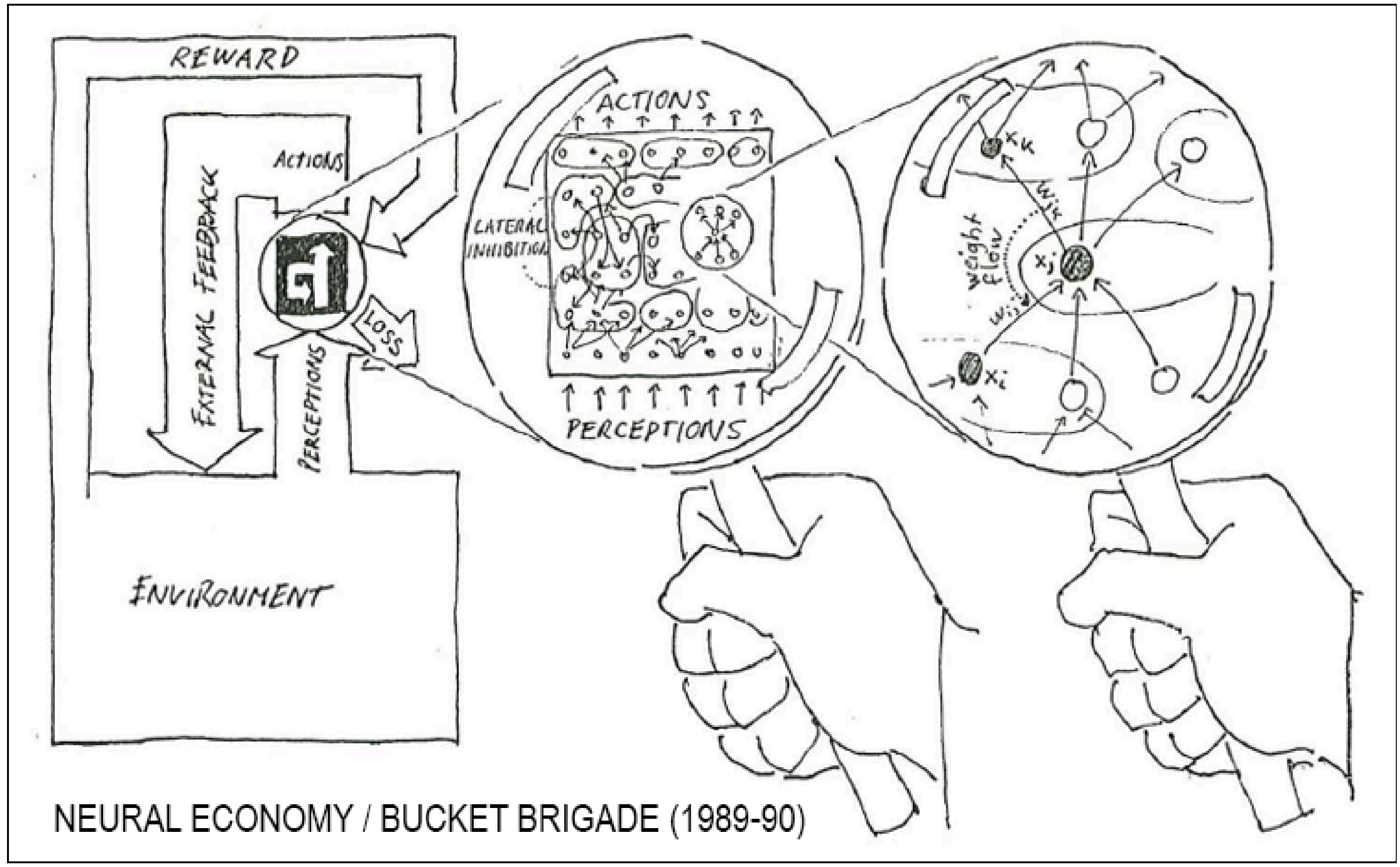

NEURAL ECONOMY / BUCKET BRIGADE (1989-90)

My doctoral dissertation at TUM[PHD] also came out in 1991, summarising some of my earlier work since 1989, including the first Reinforcement Learning (RL) *Neural Economy* (the Neural Bucket Brigade),[BB1-2][DLP] learning algorithms for RNNs that are local in space and time,[BB1] hierarchical RL (HRL) with end-to-end differentiable subgoal generators (see Sec. 10), RL and planning through a combination of two RNNs called the *controller* C and the *world model* M (see Sec. 11), sequential attention-learning NNs (see Sec. 9), NNs that learn to adjust other NNs (including "synthetic gradients;" see Sec. 15), and unsupervised or self-supervised, generative, adversarial networks (see Sec. 5) for implementing curiosity.

Back then, much of the NN research by others was inspired by statistical mechanics.[L20][I25][K41][W45][AMH1-2][NOB] The works of 1990-91 (and my even earlier diploma thesis of 1987[META1]) embodied an alternative *program-oriented* view of Machine Learning.

When Kurt Gödel laid the foundations of theoretical computer science in 1931,[GOD][GOD34][GOD21-21b] he represented both data (such as axioms and theorems) and programs (such as proof-generating sequences of operations on the data) in a *universal coding language* based on the integers. He famously used this language to construct formal statements that talk about the computation of other formal statements—especially self-referential statements which imply that they are not decidable, given a computational theorem prover that systematically enumerates all possible theorems from an enumerable set of axioms. Thus he identified fundamental limits of algorithmic theorem proving, computing, and any type of computation-based AI.

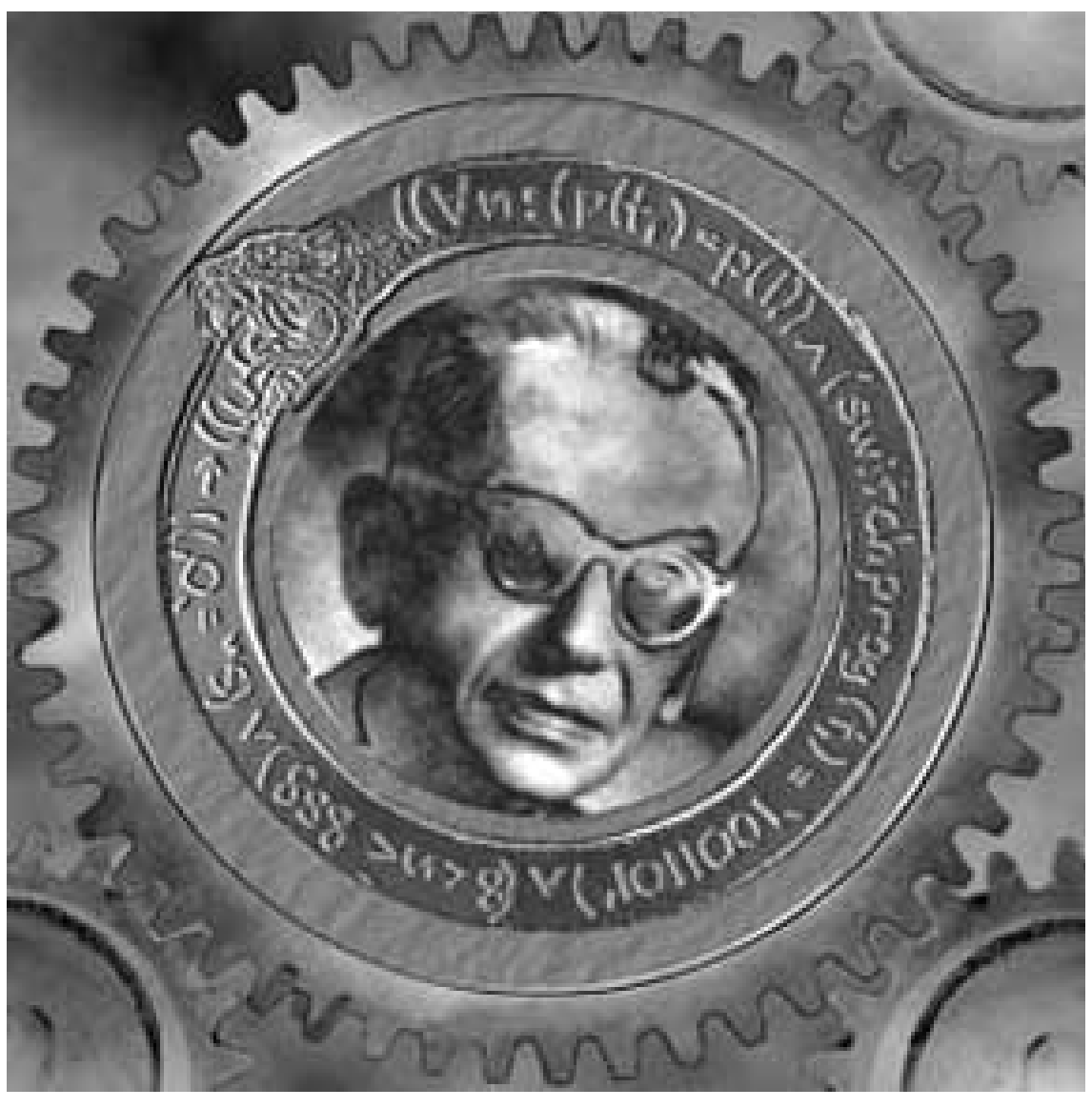

As I have frequently pointed out since 1990,[AC90] the weights of an NN should be viewed as its program. Some argue that the goal of a deep NN is to learn useful *internal representations* of observed data (there even is an international conference on learning representations called ICLR), but the NN's goal is actually to learn a *program* (the parameters) that *computes* such representations. Inspired by Gödel, I built NNs whose outputs are programs or weight matrices of other NNs (see Sec. 8), and even *self-referential RNNs* that can run and inspect their own weight change algorithms or learning algorithms (see Sec. 8). A difference to Gödel's work is that the universal programming language is not based on the integers, but on real values, such that the outputs of typical NNs are differentiable with respect to their programs. That is, a simple program generator (the efficient gradient descent procedure[BP1]) can compute a direction in program space where one may find a better program,[AC90] in particular, a *better program-generating program* (see Sec. 8). Much of my work since 1989 has exploited this fact.

## 19. From Unsupervised Pre-Training to Pure Supervised Learning (1991-95; 2006-11)

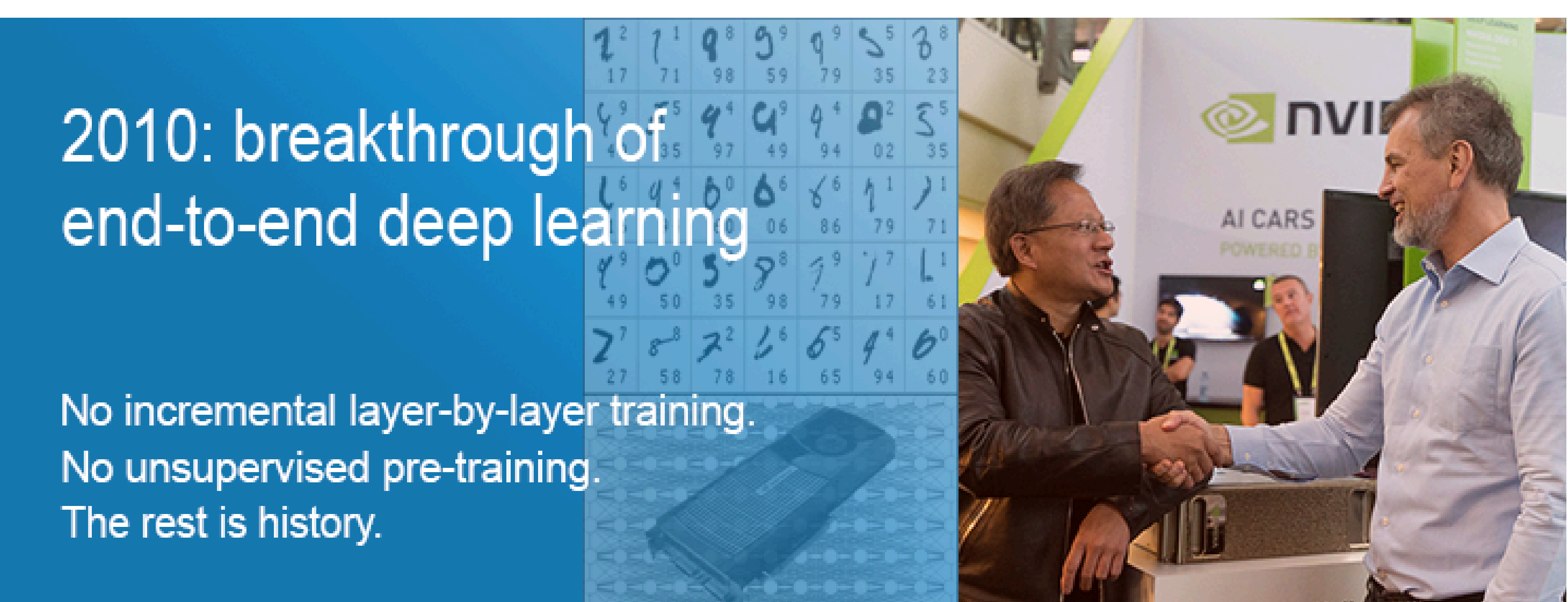

As mentioned in Sec. 1, my first Very Deep Learner was the RNN stack of 1991 which used unsupervised pre-training to learn problems of depth greater than 1000. Soon afterwards, however, we published more powerful ways of overcoming the Deep Learning Problem (see Sec. 3) *without* any unsupervised pre-training, replacing the unsupervised RNN stack[UN1-3] by the purely supervised Long Short-Term Memory (LSTM) (Sec. 4). That is, already in the previous millennium, unsupervised pre-training lost significance as LSTM did not require it. In fact, this shift from *unsupervised* pre-training to pure *supervised* learning started already in 1991.

A very similar shift took place much later between 2006 and 2010, this time for the less general *feedforward* NNs (FNNs) rather than *recurrent* NNs (RNNs). Again, my little lab played a central role in this transition. In 2006, supervised learning in FNNs was facilitated by unsupervised pre-training of stacks of FNNs[UN4] (see Sec. 1). But in 2010, our team with my outstanding Romanian postdoc Dan Ciresan[MLP1-3] showed that deep FNNs can be trained by plain backpropagation and do not at all require unsupervised pre-training for important applications.[MLP2-3] Our system set a new performance record[MLP1] on the back then famous and widely used image recognition benchmark called MNIST. This was achieved by greatly accelerating traditional FNNs on highly parallel graphics processing units called GPUs. A reviewer called this a "wake-up call to the machine learning community."

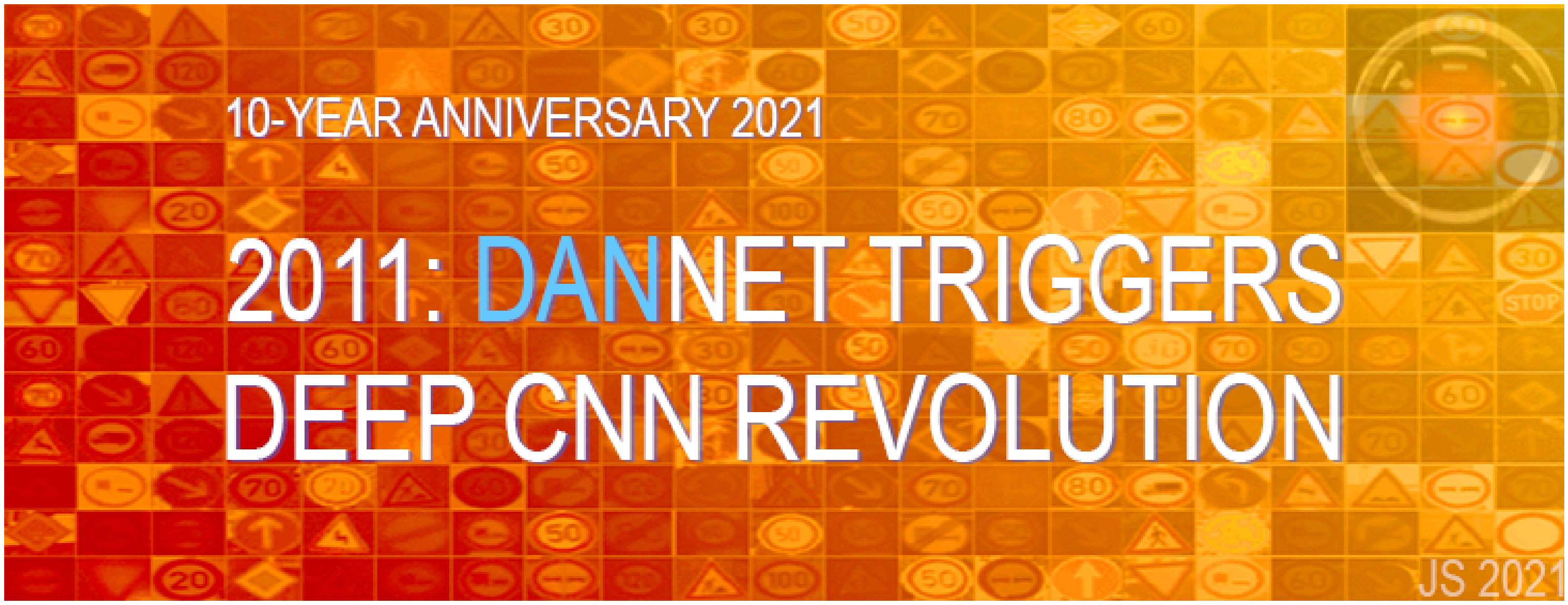

My team at the Swiss AI Lab IDSIA further improved the above-mentioned work (2010) on purely supervised Deep Learning in FNNs[MLP1-3] by replacing the traditional FNNs through another old NN type called convolutional NNs or CNNs, invented and developed by others 1979-1988 in Japan.[CN69-88] See also: Who invented convolutional neural networks?[WHO7]

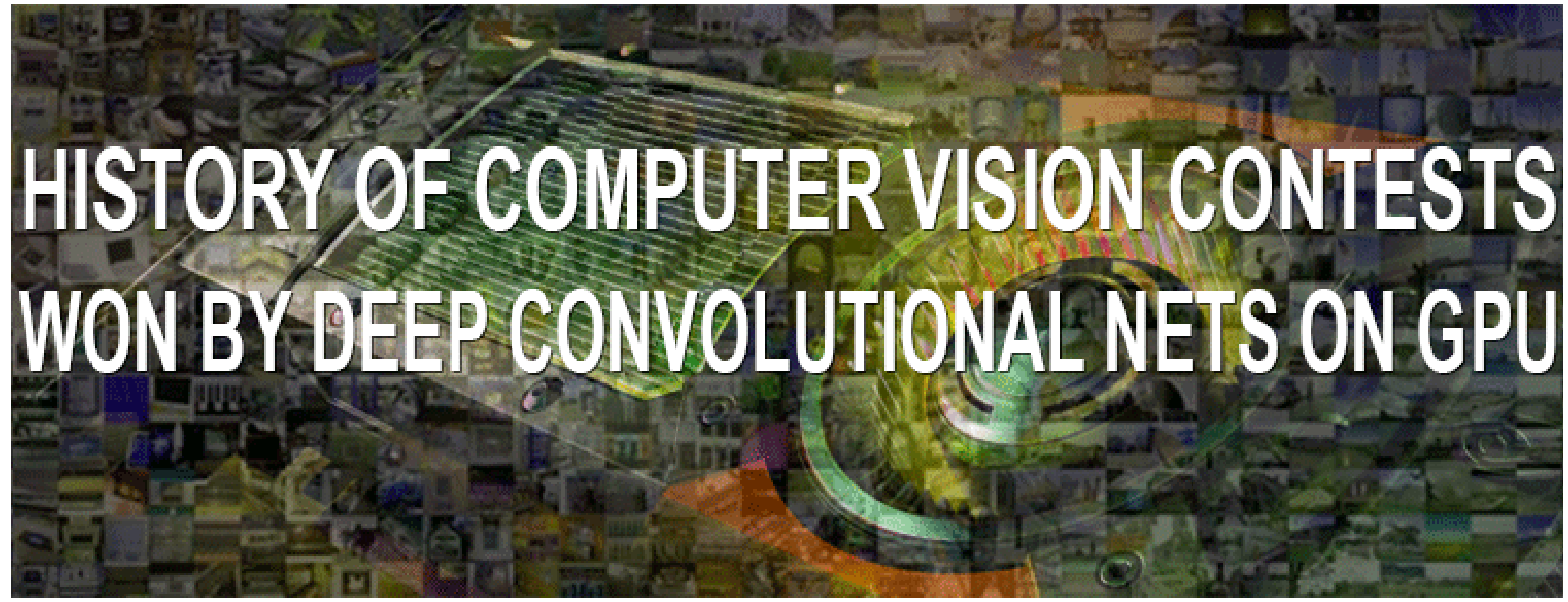

Our supervised fast deep CNN called DanNet (Ciresan et al., 2011)[GPUCNN1] was a practical breakthrough (much faster than early work on accelerating CNNs[GPUCNN]). DanNet was the first pure CNN to win international computer vision competitions, and won 4 of them in a row between May 15, 2011, and September 10, 2012.[GPUCNN5] (All of this happened before a similar GPU-CNN by others won ImageNet 2012.[GPUCNN5]) In particular, DanNet was the first deep CNN to win a Chinese handwriting contest (ICDAR 2011), the first to achieve superhuman visual pattern recognition in any international contest (IJCNN 2011), the first to win an image segmentation contest (ISBI, May 2012), and the first to win a contest on object detection in large images (ICPR, 10 Sept 2012), at the same time the first to win a medical imaging contest (on cancer detection).[GPUCNN5]

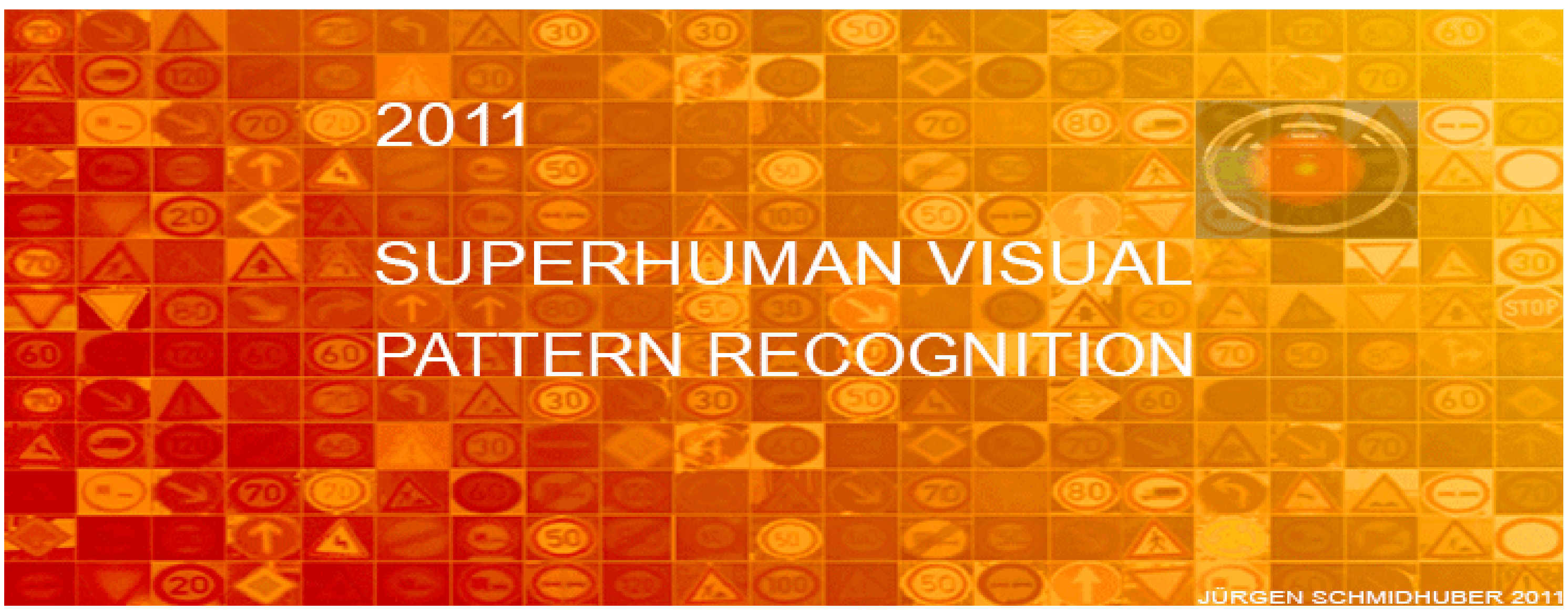

One year later, our team also won the MICCAI Grand Challenge on mitosis detection.[MGC][GPUCNN5-8] Our fast CNN image scanners were over 1000 times faster than previous methods.[SCAN] This Deep Learning approach has transformed medical imaging.

DanNet more than halved the error rate for object recognition in a contest already in 2011, 20 years after our *Annus Mirabilis*.[GPUCNN2] Soon afterwards, others applied similar approaches in image recognition contests.[GPUCNN5][MOST]

Like our LSTM results of 2009 (see Sec. 4), the above-mentioned results with *feedforward* NNs of 2010-11 attracted enormous interest from industry. For example, in 2010, we introduced our deep and fast GPU-based NNs to Arcelor Mittal, the world's largest steel maker, and were able to greatly improve steel defect detection.[ST] This may have been the first Deep Learning breakthrough in heavy industry. Today, most AI startups and major IT firms as well as many other famous companies are using such supervised fast GPU-NNs.

Let me emphasize, however, that the above-mentioned supervised deep learning revolutions of the early 1990s (for recurrent NNs) and of 2010 (for feedforward NNs)[MLP1-3] did not at all *kill* *un*supervised learning. For example, pre-trained language models are now heavily used by *Transformers* (see Sec. 8) which excel at the traditional LSTM domain of Natural Language Processing[TR1-6] (although there are still many language tasks that LSTM can rapidly learn to solve quickly[LSTM13] while plain Transformers can't). Remarkably, unnormalized linear Transformers[ULTRA] (see Sec. 8) were also first published[FWP0-2] in our Annus Mirabilis of 1990-1991,[MOST] together with unsupervised pre-training for deep learning[UN-UN3] (see the P and the T in ChatGPT). And our unsupervised generative adversarial NNs since 1990[AC90-AC20][PLAN][AC] are still used to endow agents with artificial curiosity (see Sec. 5 & Sec. 6)—see also a version of our adversarial NNs[AC90b] called GANs.[AC20][R2][PLAN][MOST][DLP] Unsupervised learning still has a bright future!

# 20. The Amazing FKI Tech Report Series on Artificial Intelligence in the 1990s

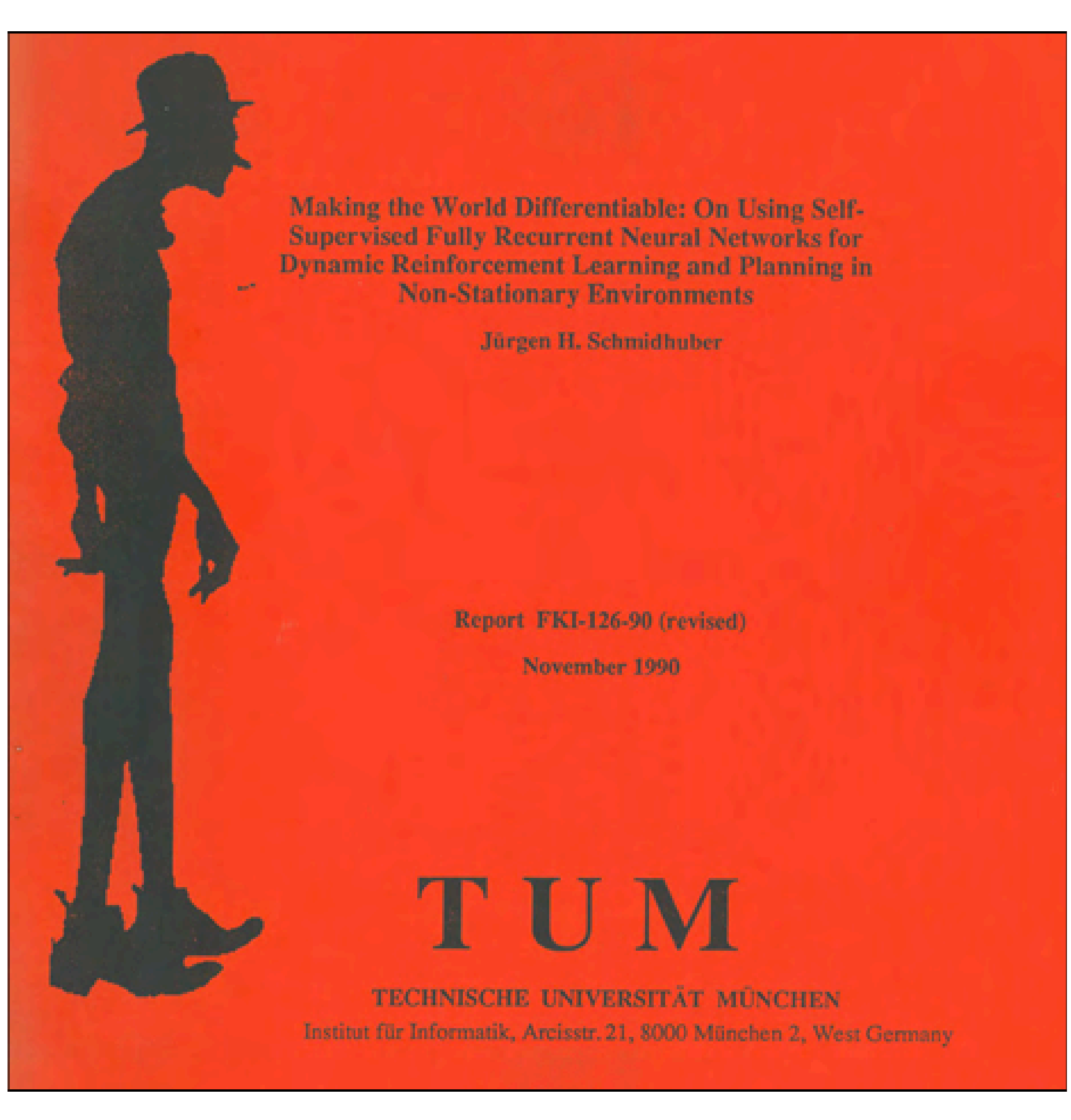

In hindsight, many of the later widely used basic ideas of "modern" Deep Learning were published in our *Miraculous Year* 1990-1991 at TU Munich, soon after the fall of the Berlin Wall: unsupervised or self-supervised, data-generating, adversarial networks (for artificial curiosity and related concepts; see Sec. 5; see also follow-up work at CU in Sec. 7), the Fundamental Deep Learning Problem (vanishing / exploding gradients; see Sec. 3) and its solutions through (a) unsupervised pre-training for very deep (recurrent) networks (see Sec. 1) and (b) basic insights leading to LSTM (see Sec. 3, Sec. 4). We also introduced sequential attention-learning NNs back then—another concept that has become popular (see Sec. 9 on both *hard* and *soft* attention, in *observation* space and in *latent* space), as well as NNs that learn to program the fast weights of another NN, and even their own weights. In particular, we already had what's now called unnormalized linear Transformers with "linearized self-attention" (see Sec. 8). Plus all the other things mentioned above, from Hierarchical Reinforcement Learning (see Sec. 10) to planning with recurrent neural world models (see Sec. 11). Of course, one had to wait for faster computers to commercialize such algorithms. By the mid 2010s, however, our stuff was massively used by Apple, Google, Facebook, Amazon, Samsung, Baidu, Microsoft, etc, many billions of times per day on billions of computers.[DL4]

Most of the results above were actually first published in TU Munich's FKI Tech Report series, for which I drew many illustrations by hand, some of them shown in the present page (see Sec. 10, Sec. 12, Sec. 8, Sec. 18). The FKI series now plays an important role in the history of Artificial Intelligence, as it introduced several important concepts: unsupervised pre-training for very Deep Learning (FKI-148-91;[UN0] see Sec. 1 and the P in ChatGPT), compressing /

distilling one NN into another (FKI-148-91;[UN0] see Sec. 2), the vanishing gradient problem and deep residual learning[WHO11] and Long Short-Term Memory (FKI-207-95;[LSTM0] see Sec. 3, Sec. 4), Artificial Curiosity through NNs that maximize learning progress (FKI-149-91;[AC91] see Sec. 6), end-to-end-differentiable Fast Weight Programmers that learn to program other NNs (separating storage and control for NNs like in traditional computers; see FKI-147-91[FWP0] and Sec. 8—the outer product version of 1991 is an unnormalized linear Transformer with linearized self-attention), learning of sequential attention with NNs (FKI-128-90;[ATT0] see Sec. 9), goal-defining commands as extra NN inputs (FKI-128-90,[ATT0] FKI-129-90;[HRL0] see Sec. 12), end-to-end-differentiable Hierarchical Reinforcement Learning (FKI-129-90;[HRL0] see Sec. 10), NNs adjusting NNs / synthetic gradients (FKI-125-90;[NAN2] see Sec. 15). (Cubic gradient computation for online recurrent NNs also was published as FKI-151-91,[CUB1] but this one does not really count; see Sec. 16.) In particular, the report FKI-126-90[AC90] introduced a whole bunch of concepts that are now widely used: planning in partially observable environments with recurrent world models (see Sec. 11), high-dimensional reward signals as extra NN inputs / general value functions (see Sec. 13), deterministic policy gradients (see Sec. 14), NNs that are both *generative* and *adversarial* (GANs; see Sec. 5; see also Sec. 7), for Artificial Curiosity and related concepts. Later remarkable FKI Tech Reports from the 1990s describe ways of greatly compressing NNs[KO0][FM] to improve their generalisation capability.

Peer-reviewed versions came out soon after the tech reports. For example, in 1992, I had a fun contest with the great David MacKay as to who'd have more publications within a single year in *Neural Computation*, back then the leading journal of our field. By the end of 1992, both of us had four. But David won, because his publications (mostly on Bayesian approaches for NNs) were much longer than mine :-) *Disclaimer:* Of course, silly measures like number of publications and h-index etc should not matter in science.[NAT1]

# 21. Concluding Remarks

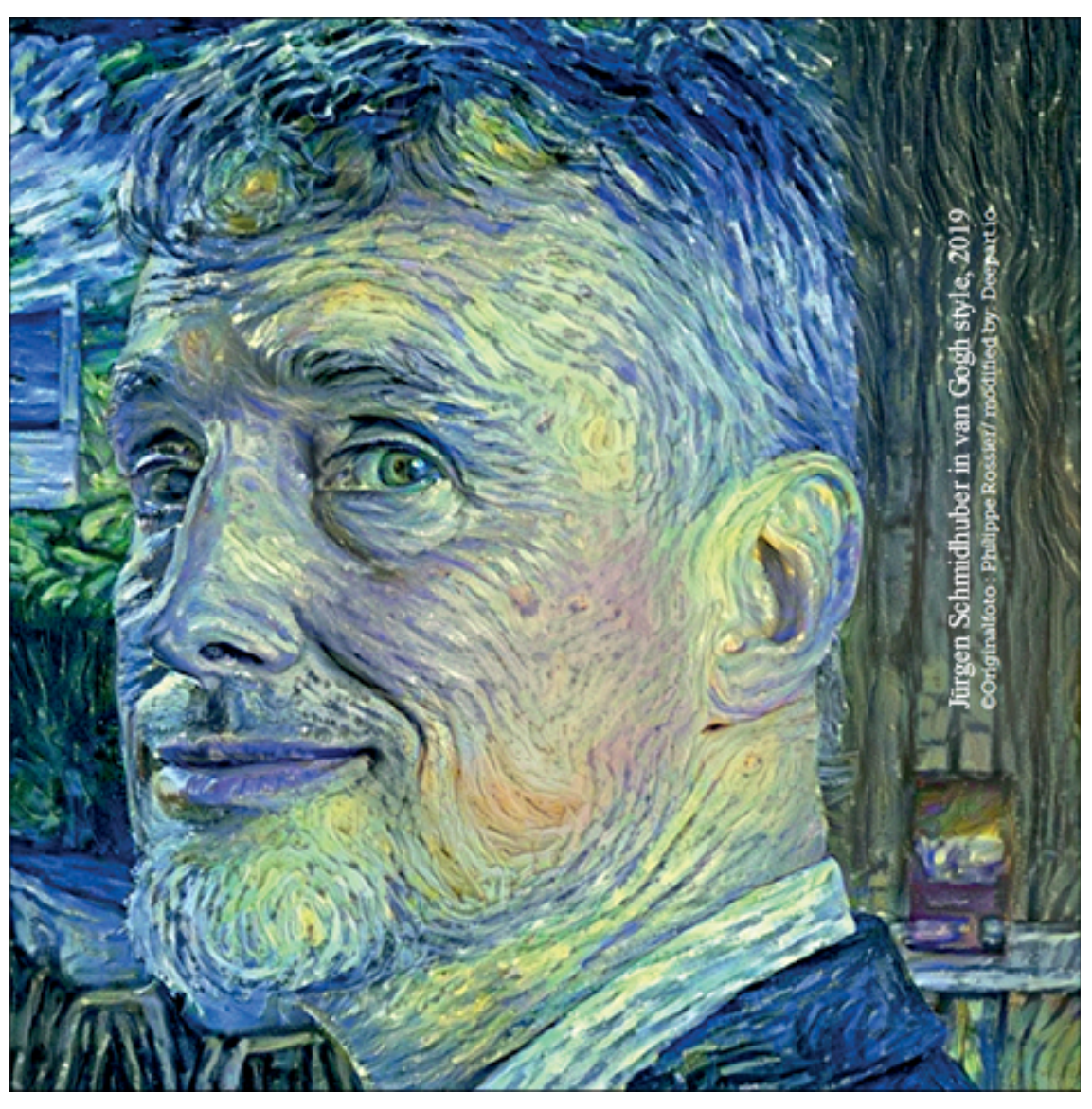

When only consulting surveys from the Anglosphere, it is not always clear[DLC] that Deep Learning was first conceived outside of it.[DLH][NOB][WHO5-11] It started in 1965 in the Ukraine (back then the USSR) with the first nets of arbitrary depth that really learned[DEEP1-2] (see Sec. 1). A few years later, stochastic gradient descent was successfully applied in Japan to learn internal representations in deep multi-layer perceptrons.[GD1-2a][DLH][DLP][NOB] Soon afterwards, modern backpropagation was published in Finland (1970)[BP1] (see Sec. 0). The basic deep convolutional NN architecture (now widely used) was invented in the 1970s in Japan[CN79][CN86] where NNs with convolutions were later (1987-88) also combined with backpropagation.[CN87][CN88][CN89]

Unsupervised or self-supervised generative adversarial networks that duel each other in a minimax game to implement artificial curiosity etc (now widely used) originated in Munich (1990, Sec. 5). So did the first self-driving cars in traffic (1994), the unnormalized linear Transformer[ULTRA][FWP,FWP0] (1991, see the T in ChatGPT & Sec. 8), and neural network distillation (1991, see DeepSeek tweet & Sec. 2). The fundamental problem of backpropagation-based Deep Learning[VAN1] (1991, see Sec. 3) was also discovered in Munich. So were the first "modern" Deep Learners to overcome this problem, through (1) unsupervised pre-training[UN1-2] (1991, see the P in ChatGPT & Sec. 1), and (2) deep residual learning[WHO11] for very deep gradient-based NNs such as Long Short-Term Memory,[LSTM0-7] "arguably the most commercial AI achievement"[AV1] (see Sec. 4). In fact, as of 2025, the two most frequently cited scientific articles of all time (with the most Google Scholar citations within 3 years—manuals excluded) are both directly based on the 1991 work at TUM.[MOST26]

LSTM was further developed in Switzerland (see Sec. 4), which is also home of the first image recognition contest-winning deep GPU-based CNNs (2011, Sec. 19—everybody in computer vision is using this approach now), the first superhuman visual pattern recognition (2011), and the first very deep, working feedforward NNs with hundreds of layers[HW1] based on deep residual learning[WHO11] (see Sec. 4). Around 1990, Switzerland also became origin of the *World Wide Web,* which allowed for quickly spreading AI around the globe. As of 2025, Switzerland is still leading the world in AI research in terms of citation impact, though since 2017, China has been the nation that produces the most papers on AI.[THE17]

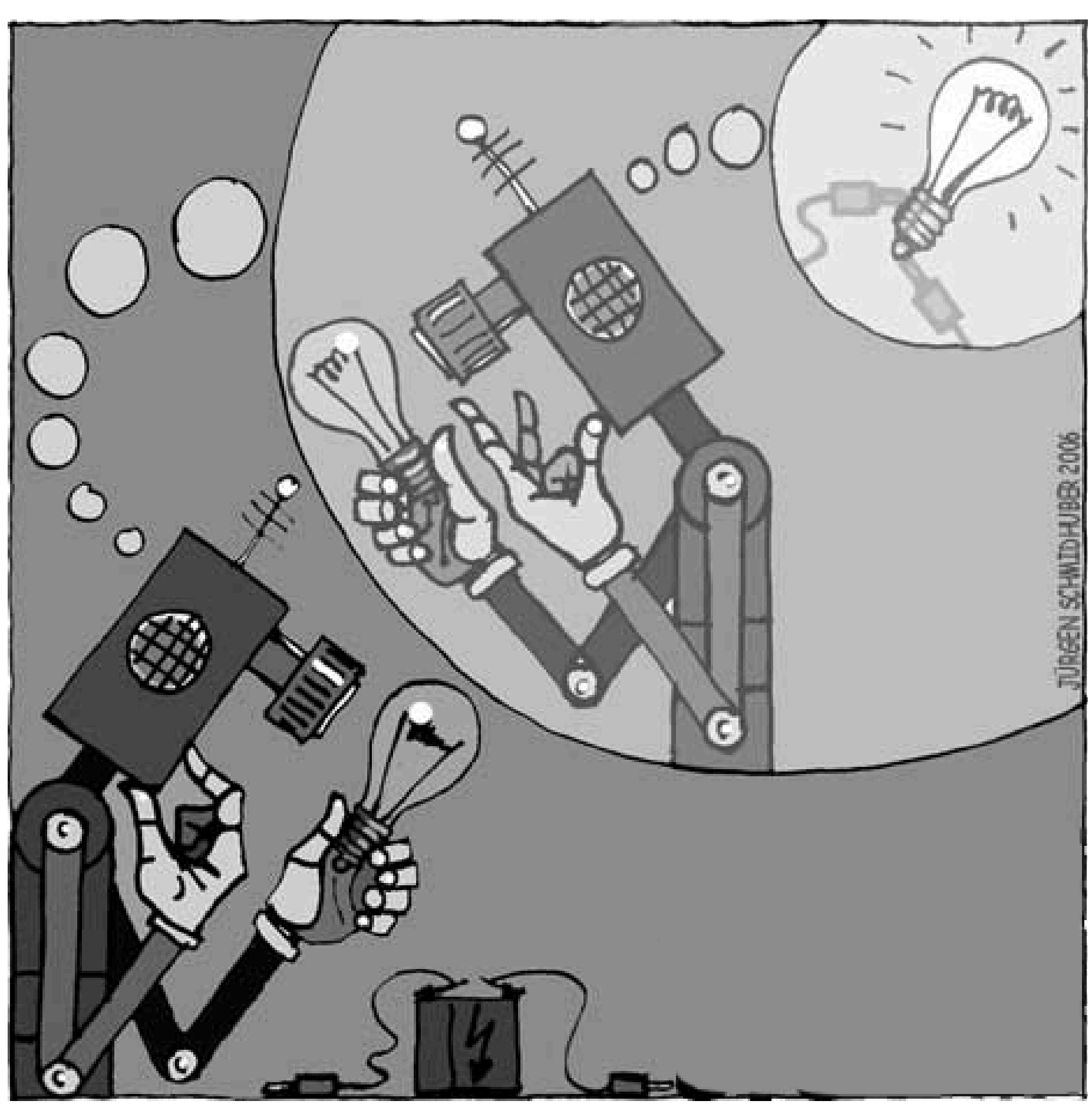

Of course, Deep Learning is just a small part of AI, mostly limited to passive pattern recognition. We view it as a by-product of our research on more general AI through meta-learning or "learning to learn learning algorithms" (publications since 1987), systems with artificial curiosity and creativity that invent their own problems and set their own goals (since 1990), evolutionary computation (since 1987) and RNN evolution and compressed network search, reinforcement learning (RL) for agents in realistic partially observable environments where traditional RL (for board games etc) does not work (since 1989), artificial super intelligence through optimal universal learning machines such as the Gödel machine (2003-), optimal search for programs running on general purpose computers such as RNNs, etc.

And of course, AI itself is just part of a grander scheme driving the universe from simple initial conditions to more and more unfathomable complexity.[SA17] Finally, even this awesome process may be just a tiny part of the even grander, optimally efficient computation of *all* logically possible universes.[ALL1-3]

# Acknowledgments

Thanks to several expert reviewers for useful comments. (Let me know under *juergen@idsia.ch* if you can spot any remaining error.) The present article[MIR] had an impact on later reports and posts which contain additional relevant references. It also influenced some of the most popular posts and comments of 2019 at reddit/ML: at the time the largest machine learning forum with over 800k subscribers.[R2-R8] The contents of this article may be used for educational and non-commercial purposes, including articles for Wikipedia and similar sites. This work is licensed under a Creative Commons Attribution-NonCommercial-ShareAlike 4.0 International License.